\documentclass[10pt,twocolumn,letterpaper]{article}

\usepackage{iccv}
\usepackage{times}
\usepackage{epsfig}
\usepackage{graphicx}
\usepackage{amsmath}
\usepackage{amssymb}
\usepackage{rotating}
\usepackage{times}
\usepackage{epsfig}
\usepackage{makecell}
\usepackage{lineno}
\usepackage{tabularx}
\usepackage{xcolor}
\usepackage{multirow}
\usepackage{graphicx}
\usepackage{hhline}
\usepackage{amsmath}
\usepackage{amssymb}
\usepackage{amsfonts}
\usepackage{empheq}
\usepackage{float}
\usepackage{framed, color}
\usepackage{xcolor}
\usepackage{graphics}
\usepackage{graphicx}
\usepackage[]{graphicx} 
\usepackage{epsfig} 
\usepackage{subfigure}
\usepackage{algorithm}
\usepackage{algorithmic}
\usepackage{stmaryrd}
\usepackage[mathscr]{eucal}
\usepackage{lineno}
\usepackage{color}
\usepackage{filecontents}
\usepackage{subfigure}
\usepackage{multirow}
\usepackage{filecontents}
\usepackage{verbatim} 
\usepackage{tabularx}
\usepackage{lineno}
\usepackage{setspace}
\usepackage{multirow}
\usepackage{textcomp,booktabs}
\usepackage{caption}

\def\D{{\bf D}}

\def\F{{\bf F}}
\def\f{{\bf f}}

\def\I{{\bf I}}

\def\X{{\bf X}}
\def\Y{{\bf Y}}

\def\U{{\bf U}}

\def\w{{\bf w}}
\def\0{{\bf 0}}
\def\1{{\bf 1}}

\def\ie{\emph{i.e. }}

\def\etal{{\em et al.\/}\,}

\newcommand*{\colorboxed}{}
\def\colorboxed#1#{%
	  \colorboxedAux{#1}%
}

\newcommand*{\colorboxedAux}[3]{%
	\begingroup
	\colorlet{cb@saved}{.}%
	\color#1{#2}%
	\boxed{%
		\color{cb@saved}%
		#3%
	}%
	\endgroup
}

\usepackage[pagebackref=true,breaklinks=true,letterpaper=true,colorlinks,bookmarks=false]{hyperref}

\iccvfinalcopy 

\def\iccvPaperID{2210} 

\ificcvfinal\fi
\begin{document}

\title{Multigrid Predictive Filter Flow for Unsupervised Learning on Videos}

\author{Shu Kong,  \ Charless Fowlkes\\
Dept. of Computer Science, University of California, Irvine\\
{\tt\small \{skong2, fowlkes\}@ics.uci.edu}
\\ \\
  \ [\href{http://www.ics.uci.edu/~skong2/mgpff.html}{Project Page}],
[\href{https://github.com/aimerykong/predictive-filter-flow}{Github}],
[\href{https://github.com/aimerykong/predictive-filter-flow/blob/master/mgPFF_video/demo01_videoSegTrack.ipynb}{Demo}],
[\href{http://www.ics.uci.edu/~skong2/slides/mgpff_public_version.pdf}{Slides}],
[\href{http://www.ics.uci.edu/~skong2/slides/mgpff_poster.pdf}{Poster}]
}

\maketitle

\begin{abstract}
   We introduce multigrid Predictive Filter Flow (mgPFF), a framework for
   unsupervised learning on videos.
   The mgPFF takes as input a pair of frames and outputs per-pixel filters
   to warp one frame to the other.
   Compared to optical flow used for warping frames,
   mgPFF is more powerful in modeling sub-pixel movement and dealing with corruption
   (e.g., motion blur).
   We develop a multigrid coarse-to-fine modeling strategy that avoids the
   requirement of learning large filters to capture large displacement.
   This allows us to train an extremely compact model (\textbf{4.6MB})
   which operates in a progressive way over multiple resolutions with shared
   weights.
   We train mgPFF on unsupervised, free-form videos and
   show that mgPFF is able to not only estimate long-range flow for frame reconstruction
   and detect video shot transitions,
   but also readily amendable for video object segmentation and pose tracking,
   where it outperforms the state-of-the-art by a notable margin
   without bells and whistles.
   Moreover,
   owing to mgPFF's nature of per-pixel filter prediction,
   we have the unique opportunity to visualize how each pixel is evolving
   during solving these tasks, thus gaining better interpretability\footnote{Due to that arxiv limits the size of files, we put high-resolution figures in the project page.}.
\end{abstract}

\section{Introduction}
\label{sec:intro}
Videos contain rich information for humans to understand the scene and
interpret the world.  However, providing detailed per-frame ground-truth labels
is challenging for large-scale video datasets, prompting work on leveraging
weak supervision such as video-level labels to learn visual features for
various tasks~\cite{abu2016youtube, kay2017kinetics, caba2015activitynet,
fouhey2018lifestyle}. Video constrained to contain primarily ego-motion has
also been leveraged for unsupervised learning of stereo, depth, odometry, and
optical flow~\cite{vijayanarasimhan2017sfm, godard2017unsupervised,
sermanet2018time, zhou2017unsupervised, zhan2018unsupervised}.

Cognitively, a newborn baby can easily track an object without understanding
any high-level semantics by watching the ambient environment for only one
month~\cite{goren1975visual, muller1978visual, moore1978visual, von2007predictive}.
However, until recently very few work has demonstrated effective unsupervised learning on
free-form videos\footnote{By ``free-form'', we emphasize the videos are long
(versus short synthetic ones~\cite{dosovitskiy2015flownet, janai2018unsupervised}), raw
and unlabeled, and do not contain
either structural pattern (e.g., ego-motion
videos~\cite{geiger2013vision, cordts2016cityscapes, yang2018every}) or
those with restricted background~\cite{tung2017self, rhodin2018unsupervised}.
}.
For example,
Wei \etal exploit the physics-inspired observation called arrow of time~\cite{popper1956arrow,
gold1962arrow}
to learn features by predicting whether frames come with the correct temporal
order~\cite{wei2018learning}, and show the features are useful in action classification
and video forensic analysis.
Vondrick \etal use video colorization as a proxy task and show that the learned features
capture objects and parts which are useful for tracking objects~\cite{vondrick2018tracking}.

\begin{figure}[t]
    \centering
    \begin{minipage}{0.49\textwidth}
        \centering
        \includegraphics[width=1\linewidth]{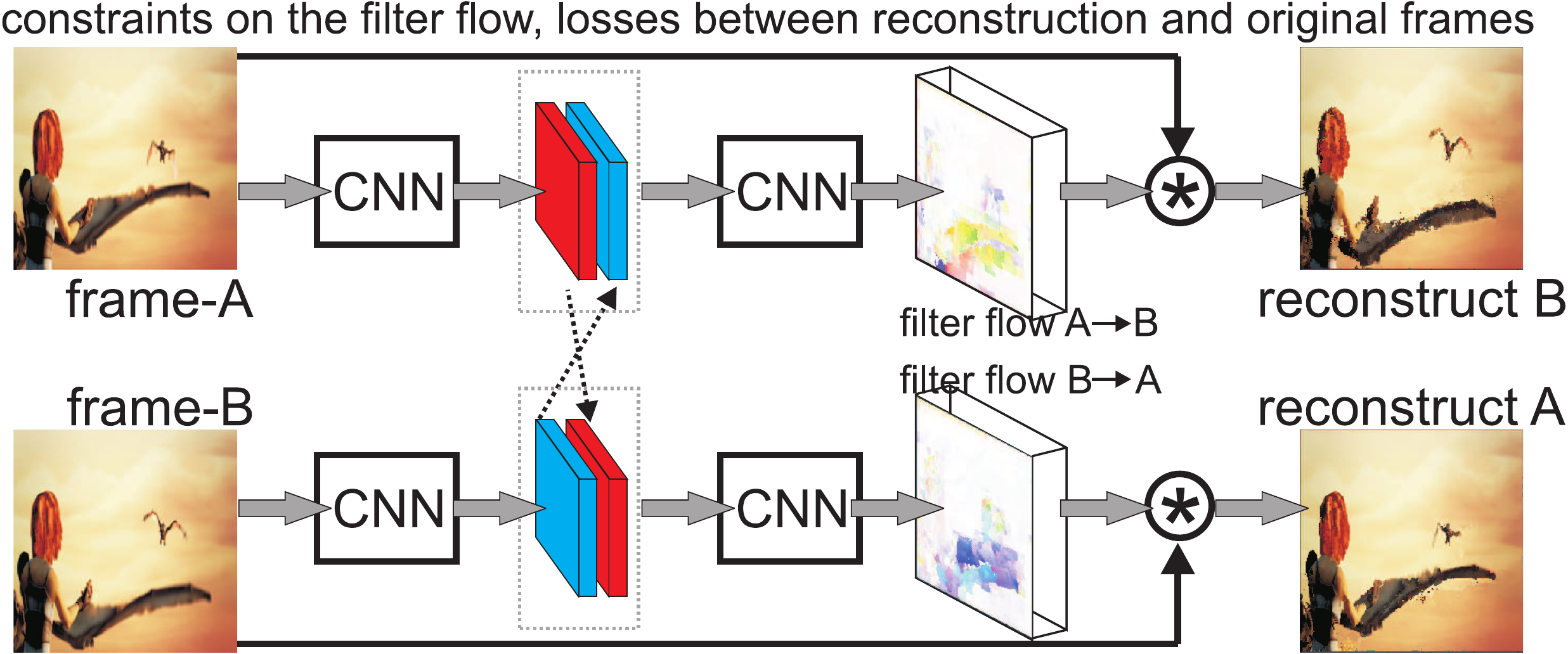}
        \captionsetup{width=0.999\textwidth}
    \end{minipage}
    \vspace{-1mm}
    \caption{\small
    The flowchart of multigrid Predictive Filter Flow framework (mgPFF).
    Conceptually we draw a single scale for demonstrating how we train our model
    in an unsupervised way with the photometric reconstruction loss along with
    constraints imposed on the filter flow maps.
    The multigrid strategy is illustrated in Fig.~\ref{fig:multigridWarping}.
    }
    \label{fig:flowchart}
    \vspace{-3mm}
\end{figure}

In this paper we explore how to train on unsupervised, free-form videos for
video object segmentation and tracking using a new framework we call multigrid
Predictive Filter Flow (mgPFF), illustrated by the conceptual flowchart in
Fig.~\ref{fig:flowchart}.  mgPFF makes direct, fine-grained predictions of how
to reconstruct a video frame from pixels in the previous frame and is trained
using simple photometric reconstruction error.  We find these pixel-level flows
are accurate enough to carry out high-level tasks such as video object
segmentation and human pose.

A straightforward approach to learning a flow between frames is to employ a
differentiable spatial transform (ST) layer (a.k.a grid
sampling)~\cite{jaderberg2015spatial}, to output per-pixel coordinate offset
for sampling pixels with bilinear interpolation and apply the transform to the
frame to estimate photometric reconstruction error.  This has been widely used
in unsupervised optical flow learning~\cite{ren2017unsupervised,
janai2018unsupervised, meister2018unflow, jason2016back, wang2018occlusion}.
However, we and others observe that unsupervised learning on free-form videos
with a simple ST-layer is challenging.  Detlefsen \etal give an excellent
explanation on why it is hard to train with ST-layer in the supervised learning
setup~\cite{skafte2018deep}.  Briefly, training with ST-layer requires the
invertibility of the spatial transform which is not guaranteed during training.
Additionally, we note that fixed grids for sampling (usually 2x2 for bilinear
interpolation) typically only provide meaningful gradients once the predicted
flow is nearly correct (i.e., within 1 pixel of the correct flow). This
necessitates training at a coarse scale first to provide a good initialization
and avoid getting caught in bad local-minima.

Inspired by the conceptual framework Filter Flow~\cite{seitz2009filter}, we
propose to learn in the mgPFF framework per-pixel filters instead of the
per-pixel offset as in the ST-layer. For each output pixel, we predict the
weights of a filter kernel that when applied to the input frame reconstruct
the output.  Conceptually, we reverse the order of operations from the
ST-layer.  Rather than predicting an offset and then constructing filter
weights (via bilinear interpolation), we directly predict filter weights
which can vote for the offset vector. We observe that training this model is
substantially easier since we get useful gradient information for all possible
flow vectors rather than just those near the current prediction.

Since the filter-flow approach outputs per-pixel kernels during training,
capturing large displacements is computationally expensive.  We address this
using a multigrid strategy~\cite{trottenberg2000multigrid, hackbusch2013multi}
to approximate the kernel.  Concretely, we run the model over
multi-resolution inputs with a fixed filter size (11x11 used in this paper) and
compose the filters generated at multiple scales to produce the final flow
fields (detailed in Section \ref{ssec:mgPFF} and illustrated by
Fig.~\ref{fig:multigridWarping}). The model thus only outputs 11*11=121
per-pixel filter weights at each resolution scale (smaller than the channel
dimension in modern CNN architectures). We further assume self-similarity
across scales and learn only a single set of shared learned model weights.
This makes our model quite efficient w.r.t running time and model size.  As a
result, our final (un-optimized) model is only \textbf{4.6MB} in size and takes
0.1 seconds to process a pair of 256x256-pixel resolution images.

To summarize our contributions:
(1) conceptually,
we introduce a simple multigrid Predictive Filter Flow (mgPFF) framework allowing for
unsupervised learning on free-form videos;
(2) technically,
we show the filter flow overcomes the limitation of spatial-transform layer and the multigrid
strategy significantly reduces model size;
(3) practically,
we show through experiments that mgPFF substantially outperforms other
state-of-the-art applications of unsupervised flow learning on challenging
tasks including video object segmentation, human pose tracking and long-range
flow prediction.

\section{Related Work}

\noindent {\bf Unsupervised Learning for Vision:}
Our work builds upon a flurry of recent work
that trains visual models without human supervision.
A common approach is to leverage the natural context in images
and video for learning the visual
features~\cite{doersch2015unsupervised, owens2016ambient, jayaraman2015learning,
doersch2017multi, wang2017transitive, zhang2017split, larsson2017colorization,
pathak2016context, wang2015unsupervised, vondrick2016generating, noroozi2016unsupervised,
pathak2017learning},
which can be transferred to down-stream tasks,
such as object detection. Other approaches include
interaction with an environment to learn visual features~\cite{pinto2016curious,
agrawal2016learning, wu2016physics}, which is useful
for applications in robotics.  A related but different line of work explores
how to learn geometric properties or cycle consistencies with self-supervision,
for example for motion capture or correspondence~\cite{tung2017self,
zhou2017unsupervised, zhou2016learning, ilg2017flownet, zhou2016view, xiaolong2019learning}.
Ours also develop an unsupervised model,
but with the signal from temporal consistency
between consecutive frames in free-form videos,
without the requirement of synthetic data~\cite{zhou2016learning,ilg2017flownet}.

\noindent {\bf Unsupervised Learning on Free-Form Videos:}
Though there are a lot methods for unsupervised optical flow
learning~\cite{meister2018unflow, wang2018occlusion}
on videos (either synthetic~\cite{dosovitskiy2015flownet, janai2018unsupervised}
or structured~\cite{cordts2016cityscapes, geiger2013vision}),
there is very few work about unsupervised learning on free-form videos:
\cite{wang2015unsupervised} uses an offline tracker to provide signal to guide feature learning;
\cite{wei2018learning, misra2016shuffle, fernando2017self} learn to verify whether
frames come with the correct order, and transfer the feature to action classification;
\cite{pathak2017learning} learns for region segmentation on image by considering the moving pattern
of rigid objects;
\cite{vondrick2018tracking} learns for video colorization and shows that the learned features
capture object or parts which are useful for object tracking;
\cite{xiaolong2019learning} learns correspondence at patch level on videos with
reconstruction between frames.

\noindent{\bf Filter Flow}~\cite{seitz2009filter} is a powerful
framework which models a wide range of low-level vision problems as estimating
a spatially varying linear filter. This includes tasks such as optical flow~\cite{revaud2015epicflow,
menze2015discrete, yu2015direct},
deconvolution~\cite{levin2009understanding, perrone2016clearer, hirsch2010efficient},
non-rigid morphing~\cite{newcombe2015dynamicfusion},
stereo~\cite{scharstein2002taxonomy,mei2013segment}
defocus~\cite{li2013bayesian},
affine alignment~\cite{lazebnik2004semi},
blur removal~\cite{hirsch2011fast}, etc.
However, as it requires an optimization-based solver,
it is very computationally expensive, requiring  several hours to
compute filters for a pair of medium-size images~\cite{seitz2009filter, ravi2017filter}.
Kong and Fowlkes propose Predictive Filter Flow, which learns to predict per-pixel filters
with a CNN conditioned on a single input image to
solve various low-level image reconstruction
tasks~\cite{kong2018image}.
There are other methods embracing the idea of predicting per-pixel filters,
e.g., \cite{mildenhall2018burst} and
\cite{niklaus2017videoAdaptiveConv} do so for solving burst
denoising and video frame interpolation, respectively.

\section{Multigrid Predictive Filter Flow}
\label{sec:mgPFF}
Our multigrid Predictive Filter Flow (mgPFF) is rooted in
the Filter Flow framework~\cite{seitz2009filter},
which models the image transformations $\I_B \rightarrow \I_A$ as a linear
mapping where each pixel in $\I_A$ only depends on the local neighborhood centered at
same place in $\I_B$.
Finding such a flow of per-pixel filter
can be framed as solving a constrained linear system
\begin{equation}
\I_A = {\bf T}_{B \rightarrow A}  \cdot \I_B, \quad {\bf T}_{B \rightarrow A} \in {\bf\Gamma}.
\label{eq:filterflow}
\end{equation}
where ${\bf T}_{B \rightarrow A}$ is a matrix whose rows act separately
on a vectorized version of the source image $\I_B$.
${\bf T}_{B \rightarrow A} \in {\bf\Gamma}$ serves as a placeholder for the entire set of
additional constraints on the operator which enables a unique solution
that satisfies our expectations for particular problems of interest.
For example, standard convolution corresponds to
${\bf T}_{B \rightarrow A}$ being a circulant
matrix whose rows are cyclic permutations of a single set of filter weights
which are typically constrained to have compact localized non-zero support.
For a theoretical perspective, Filter Flow model~\ref{eq:filterflow} is simple
and elegant, but directly solving Eq.~\ref{eq:filterflow} is intractable for
image sizes we typically encounter in practice, particularly when the filters
are allowed to vary spatially.

\subsection{Predictive Filter Flow (PFF) on Video}
\label{ssec:PFF}
Instead of optimizing over $\bf T$,
Kong and Fowlkes propose the Predictive Filter Flow (PFF)
framework that learns function $f_{\w}(\cdot)$ parameterized by $\w$
that predicts the transformation $\bf T$ specific to image $\I_B$
taken as input~\cite{kong2018image}:
\begin{equation}
\I_A \approx { \bf T}_{B \rightarrow A} \cdot \I_B,
\quad {\bf {T}}_{B \rightarrow A}\equiv f_{\w}(\I_B),
\label{eq:predictivefilterflow}
\end{equation}
The function $f_{\w}(\cdot)$ is learned with a CNN model under the assumption
that $(\I_A,\I_B)$ are drawn from some fixed joint distribution.
Therefore,
given sampled image pairs, $\{(\I_A^i, \I_B^i)\}$, where $i=1,\dots, N$,
we can learn parameters $\w$ that minimize the difference between a recovered image
$\hat\I_A$ and the real one $\I_A$ measured by some loss $\ell$.

In this work,
to tailor the PFF idea to unsupervised learning on videos,
under the same assumption that $(\I_A,\I_B)$ are drawn from some fixed joint
distribution,
we can have the predictable transform
${\bf {T}}_{B \rightarrow A}\equiv f_{\w}(\I_B, \I_A)$,
parametrized by $\w$.
To learn the function $f_{\w}(\cdot)$,
we use the Charbonnier function~\cite{BruhnW05} to measure the pixel-level
reconstruction error,
defined as $\phi(s)=\sqrt{s^2+0.001^2}$,
and learn parameters $\w$ by minimizing the following objective function:
\begin{equation}
\begin{split}
&\ell_{rec}(\I_B,\I_A) = \phi( \I_A - {\bf {T}}_{B \rightarrow A} \cdot\I_B ), \\
\end{split}
\label{eq:PFF}
\end{equation}
Note that the above loss can take image pairs in different order simply by concatenating
the pixel embedding features from the two frames one over another,
as demonstrated in Fig.~\ref{fig:flowchart}.
After concatenation,
we train a few more layers to produce the per-pixel filters.

Also note that,
when exploiting the locality constraints (similar to convolution),
we implement the operation ${\bf {T}}_{B \rightarrow A} \cdot\I_B$
with the ``im2col'' function which vectorizes the local neighborhood
patch centered at each pixel
and computes the inner product of this vector with the corresponding
predicted filter.  Note that ``im2col'' and the follow-up inner product
are highly optimized for available hardware architectures in most deep
learning libraries, \emph{exactly the same} used in modern convolution operation;
thus our model is quite efficient in computation.

\subsection{Multigrid PFF}
\label{ssec:mgPFF}

\begin{figure}[t]
    \centering
    \begin{minipage}{0.45\textwidth}
        \centering
        \includegraphics[width=1\linewidth]{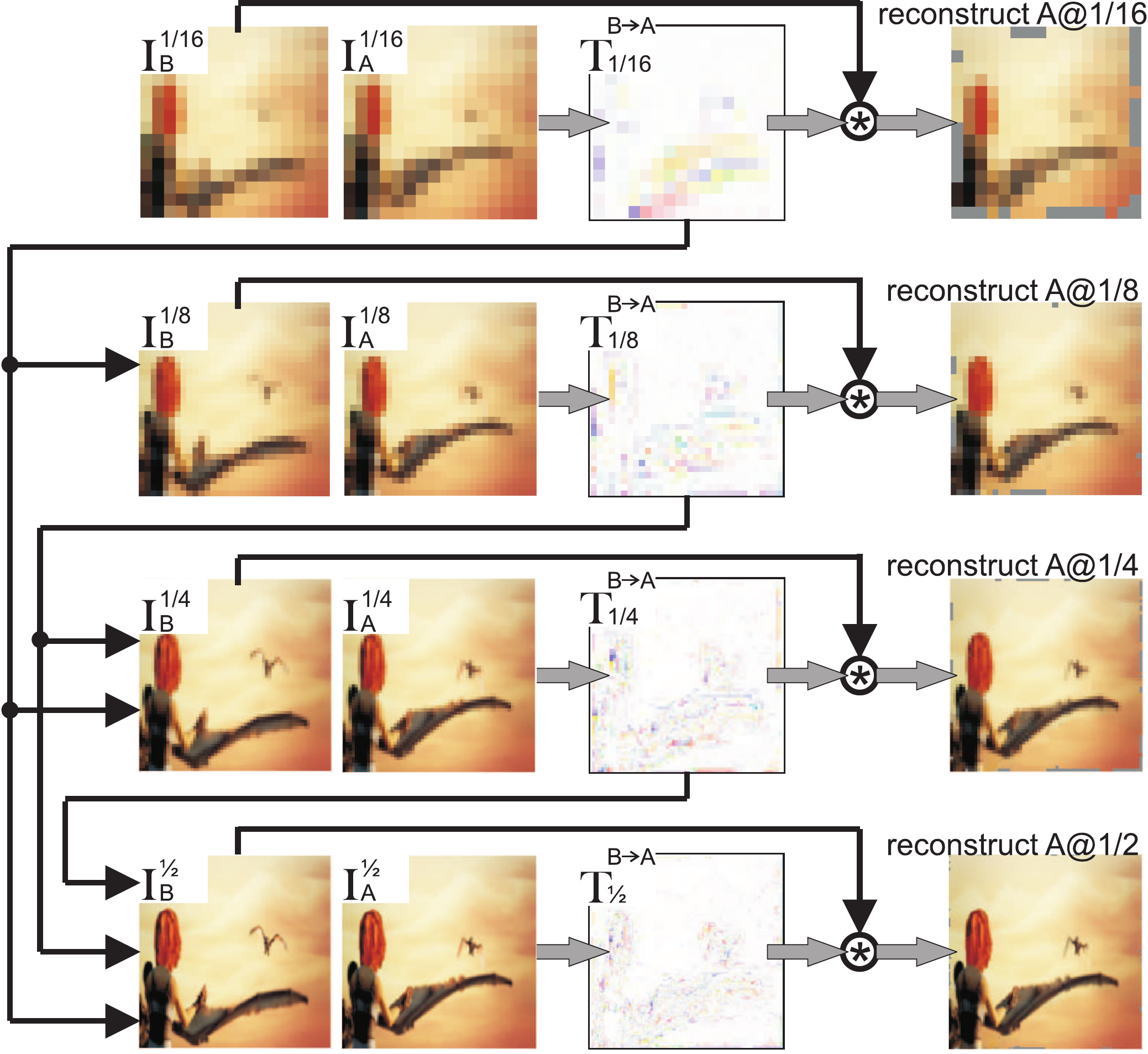}
        \captionsetup{width=0.999\textwidth}
    \end{minipage}
    \vspace{-1mm}
    \caption{\small Illustration of how multigrid Predictive Filter Flow (mgPFF)
    performs progressively by warping images from one to the other at multiple
    resolution scales from coarse to fine.
    After the finest scale,
    one can accumulate all the intermediate filter flow maps for the final one,
    which can be either transformed into optical flow or
    used for video segmentation and tracking.
    }
    \label{fig:multigridWarping}
    \vspace{-3mm}
\end{figure}

While the PFF described above is elegant and simple for unsupervised learning over videos,
it faces the substantial challenge that, to capture large displacement,
one must predict per-pixel filters with very large spatial support.
To address this problem, we are inspired by the multigrid strategy which seeks
to solve high-dimensional systems of equations using hierarchical, multiscale
discretizations of linear operators~\cite{trottenberg2000multigrid,hackbusch2013multi},
to produce a coarse-to-fine series of smaller, more easily solved problems.

To explain this, mathematically,
suppose we have filter flow ${\bf T}$ in original resolution that maps from $\X$ to
$\Y$, \ie $\Y =  {\bf T} \cdot \X$.
Then if we downsample $\X$ and $\Y$ by half,
we have
\begin{equation}
\begin{split}
\D_{\frac{1}{2}}\Y = & \D_{\frac{1}{2}}{\bf T} \cdot \X \approx (\D_{\frac{1}{2}}{\bf T}) \cdot (\U_{2\times}\D_{\frac{1}{2}}\X),\\
\end{split}
\end{equation}
where the upsampling $\U_{2\times}$ and downsampling $\D_{\frac{1}{2}}$ operators are
approximately inverse to each other.
Then we write a reduced system:
\begin{equation}
\begin{split}
\Y_{\frac{1}{2}} &  \approx (\D_{\frac{1}{2}} {\bf T} \U_{2\times})\cdot (\D_{\frac{1}{2}}\X)
= {\bf T}_{\frac{1}{2}} \X_{\frac{1}{2}}\\
\end{split}
\end{equation}
The above derivation implies we can solve a smaller system for ${\bf
T}_{\frac{1}{2}}$ on the input $\X_{\frac{1}{2}}$, e.g., an image with half the
resolution and then upsample ${\bf T}_{\frac{1}{2}}$ to get an approximate
solution to the original problem.

In practice, to avoid assembling the full resolution ${\bf T}$, we always
represent it as a composition of residual transformations at each scale.  ${\bf
T} = {\bf T}_{1} \cdot \U_{2\times} \cdot {\bf T}_{\frac{1}{2}}  \dots \U_{2\times} \cdot {\bf T}_{\frac{1}{2^{L}}}$,
where ${\bf T}_{\frac{1}{2^l}}$ is estimated filter flow over frames at
resolution scale $1/2^{l-1}$.  In our work, we set $L$=$5$.  Each individual
transformation has a fixed filter support (sparse).  By construction, the
effective filter ``sizes'' grow spatially larger as it goes up in the pyramid
but the same filter weight is simply applied to larger area (we use
nearest-neighbor upsampling).  Then the total number of filter coefficients to
be predicted for the pyramid would be just 4/3 more than just the finest level
(ref. geometric series
$4/3=1+\frac{1}{2^2}+\frac{1}{4^2}+\frac{1}{8^2}+\dots$).

Concretely,
suppose we need the kernel size as 80x80 to capture large displacement,
we can work on coarse scale of 8x smaller input region with kernel size 11x11,
this will reflect on the original image of receptive field as large as 88x88.
But merely working on such coarse scale introduces checkerboard effect if we resize
the filters 8x larger.
Therefore,
we let the model progressively generate a series of 11x11 filters
at smaller scales of [8x,4x,2x,1x], as demonstrated by Fig.~\ref{fig:multigridWarping}.
Finally,
we can accumulate all the generated filter flows towards the single map,
which can be a long-range flow (studied in Section~\ref{ssec:long-range-flow}).
We train our system with the same model at all these scales.
We have also trained scale-specific models,
but we do not observe any obvious improvements in our
experiments.
We conjecture that in diverse, free-form videos there is substantial
self-similarity in the (residual) flow across scales.

We note that coarse-to-fine estimation of residual motion is a classic approach
to estimating optical flow (see, e.g.~\cite{fleet2006optical}). It has also
been used to handle problems of temporal aliasing~\cite{simoncelli199914} and
as a technique for imposing a prior smoothness
constraint~\cite{szeliski2012bayesian}. Framing flow as a linear operator
draws a close connection to multigrid methods in numerical
analysis~\cite{trottenberg2000multigrid,hackbusch2013multi}. However, in
literature there is primarily focused on solving for ${\bf X}$ where the
residuals are additive, rather than ${\bf T}$ where the residuals are
naturally multiplicative.

\subsection{Imposing Constraints and Training Loss}
\label{ssec:loss_constraints}

We note that training with above reconstruction loss alone gives very good
reconstruction performance, but we need other constraints to regularize
training to make it work on video segmentation and tracking.  Now we describe
useful constraints used in this work.

\noindent{\bf Non-negativity and Sum-to-One}
With the PFF framework, it is straightforward to impose the non-negativity and
sum-to-one constraints by using the softmax layer to output the per-pixel
filters, as softmax operation on the kernels naturally provides a
transformation on the weights into the range of [0,1], and sum-to-one
constraint mimics the brightness constancy assumption of optical flow.

\noindent{\bf Warping with Flow Vector}
In order to encourage the estimated filter kernels to behave like optical
flow (i.e., a translated delta function) we define a projection of the
filter weights on to the best approximate flow vector by treating the
(positive) weights as a distribution and computing an expectation.
Given a filter flow ${\bf T}$ we define the nearest optical flow as
\begin{equation}
\F({\bf T}) \equiv
    \begin{bmatrix}
    v_x(i,j)\\ v_y(i,j)
    \end{bmatrix}
= \sum_{x,y} {T}_{ij,xy}
    \begin{bmatrix}
    x - i\\ y - j
    \end{bmatrix}
\label{eq:ff2of}
\end{equation}

As discussed in Section~\ref{sec:intro}, directly learning to predict $\F$ is
difficult but when keep ${\bf T}$ as an intermediate representation, learning
becomes much easier. To encourage predicted ${\bf T}$ towards a unimodal
offset, we add a loss term based on the optical flow $\F$ with grid sampling
layer just as done in literature of unsupervised optical flow learning
~\cite{ren2017unsupervised, wang2018occlusion, janai2018unsupervised}.  We
denote the loss terms as $\ell_{flow}(\I_B,\I_A)$ meaning the reconstruction
loss computed by warping with optical flow $\F({\bf T}_{B \rightarrow A})$ from
$\I_B$ to $\I_A$.

\noindent{\bf Forward-Backward Flow Consistency}
As we know,
there are many solutions to the reconstruction problem.
To constrain this for more robust learning,
we adopt a forward-backward consistency constraint as below:
\begin{equation}
\begin{split}
&\ell_{fb}({\bf f},{\bf b}) \equiv \frac{1}{\vert \cal{I} \vert}
\sum_{{\bf i} \in \cal{I}}
\phi( {\bf p}_i - {\bf b} (\f ( {\bf p}_i)) ) \\
\end{split}
\label{eq:FowardBackwardConsistency}
\end{equation}
where forward and backward flow are ${\bf f}=\F({\bf T}_{B \rightarrow A})$
and ${\bf b} = \F({\bf T}_{A \rightarrow B})$,
and ${\bf p}_i \equiv[x_i,y_i]^{T}$ is the spatial coordinate.
We note that such constraint is useful for addressing the chicken-and-egg problem related
to optical flow and occlusion/disocclusion~\cite{hur2017mirrorflow, meister2018unflow}.
But here we do not threshold the consistency error to find occlusion regions or ignore
the errors in the region.
We note that it is crucial to train the mgPFF model with this constraint
when applying the model later for video segmentation and tracking;
otherwise pixels in the object would diffuse to the background easily.

\noindent{\bf Smoothness and Sparsity}
Smoothness constraints can be done easily using traditional penalties on the norm of the
flow field gradient, \ie $\ell_{sm}\equiv \Vert \triangledown \F({\bf T}) \Vert_1$.
The smoothness penalty helps avoid big transitions on flow field,
especially at coarse scales where very few big flows are expected.
The sparsity constraint is imposed on the flow field as well with L1 norm,
\ie $\ell_{sp} \equiv \Vert \F({\bf T}) \Vert_1$.
This forces the model not to output too many abrupt flows especially at
finer scales.

Our overall loss for training mgPFF model minimizes the following
combination of the terms across multiple scales $l=1\dots,L$:
\begin{equation}
\small
\begin{split}
\min_{\w}
\sum_{l=1}^{L} & \ell_{rec}(\I_B^{l},\I_A^{l}) +
\lambda_{fl} \cdot \ell_{fl}(\I_B^{l},\I_A^{l}) \\
+ \lambda_{fb}\cdot & \ell_{fb}({\bf f}^{l},{\bf b}^{l})
+ \lambda_{sm}\cdot\ell_{sm}({\bf f}^{l})
+ \lambda_{sp}\cdot\ell_{sp} ({\bf f}^{l})\\
{\text s.t. \ \ \ \ } & {\bf {T}}^l_{B \rightarrow A}=f_{\w}(\I_B^{l},\I_A^{l}),
 {\bf {T}}^l_{A \rightarrow B} = f_{\w}(\I_A^{l},\I_B^{l}), \\
& {\bf f}^l=\F({\bf {T}}^l_{B \rightarrow A}), {\bf b}^l=\F({\bf {T}}^l_{A \rightarrow B}).
\end{split}
\label{eq:mgPFF}
\end{equation}
For simplicity,
we only write the losses involving flow from $B$ to $A$; in practice,
we also include those $A$ to $B$.

\subsection{Implementation and Training}
Our basic framework is largely agnostic to the choice of architectures.
In this paper,
we modify the ResNet18~\cite{he2016deep}
by removing  res$_4$ and res$_5$ (the top 9 residual blocks, see appendix)
and reducing the unique channel size
from $[64, 128, 256, 512]$ to $[32, 64, 128, 196]$.
We also add in bilateral connection and upsampling layers to make it a
U-shape~\cite{ronneberger2015u}, whose output is at the original resolution.
Furthermore,
we build another shallow stream but in full-resolution manner with
batch normalization~\cite{ioffe2015batch} between a convolution
layer and ReLU layer~\cite{nair2010rectified} that learns to
take care of aliasing effect caused by pooling layers in the first steam.
We note that
our mgPFF is very compact that the overall model size is only \textbf{4.6MB};
it also performs fast that the wall-clock time for processing a pair of 256x256
frames is 0.1 seconds.
Two-stream architecture is popular in multiple domain learning~\cite{simonyan2014two},
but we note that such design on a single domain was first used in \cite{pohlen2017full}
which is more computationally
expensive that the two streams talk to each other along the whole network flow; whereas ours is cheaper that
they only talk at the top layer.
We note that our architecture is different from FlowNetS and FlowNetC~\cite{dosovitskiy2015flownet}
in that,
1) unlike FlowNetS, ours produces pixel embedding features for each of frame,
which can be potentially transferred to other tasks
(though we did not explore this under the scope of this paper);
2) unlike FlowNetC, ours does not exploit the computationally expensive correlation layer.


\begin{figure}[t]
    \centering
    \begin{minipage}{0.4\textwidth}
        \centering
        \includegraphics[width=1\linewidth]{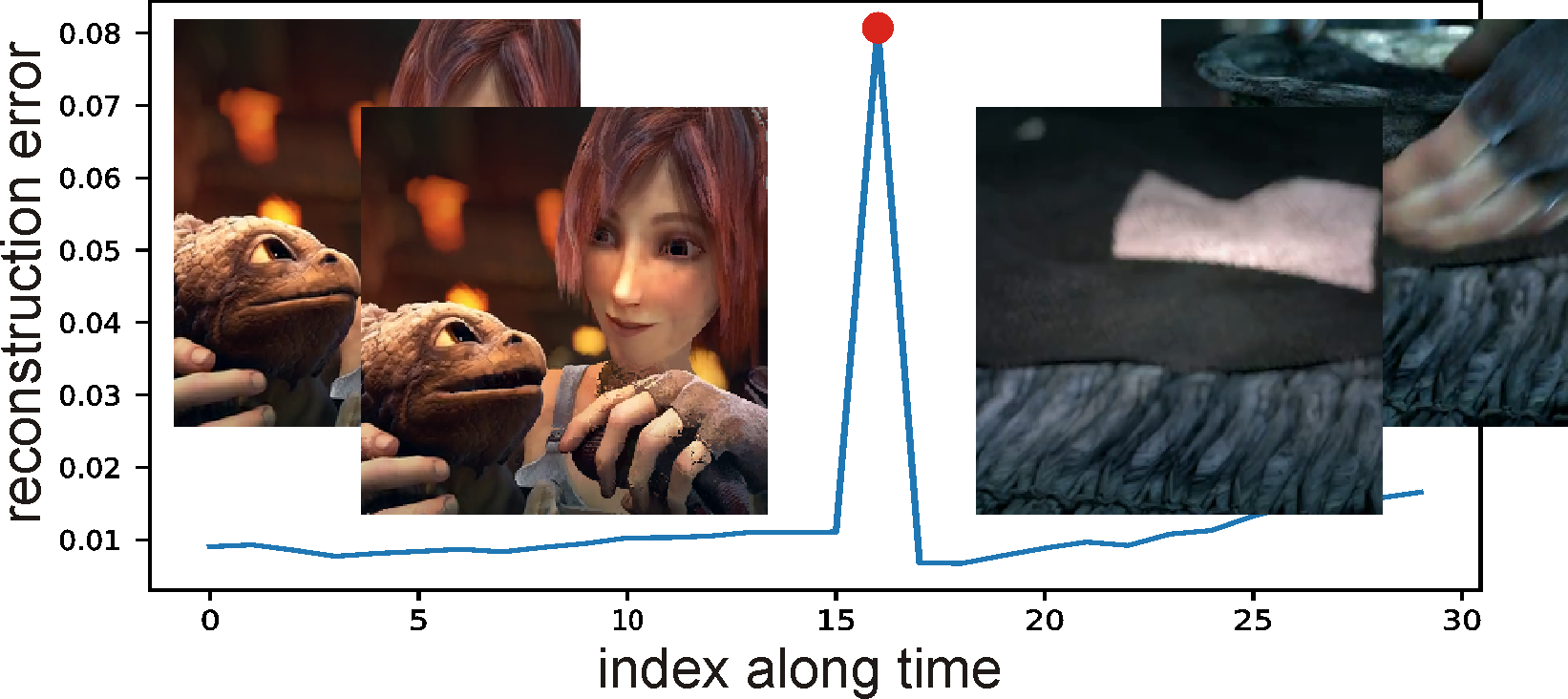}
        \captionsetup{width=0.999\textwidth}
    \end{minipage}
    \vspace{-4mm}
    \caption{\small {\bf Shot Detection arising from training on free-form videos}:
     By training our mgPFF on the Sintel movie,
     we can detect the transition shot purely based on the reconstruction error.
     This helps develop a stage-wise training that we train mgPFF first on the whole movie,
     and then simply threshold the reconstruction errors for shot detection and get
     discrete groups for finer training.
    }
    \label{fig:shotDet}
    \vspace{-3mm}
\end{figure}

\section{Experiments}
We conduct experiments to show the mgPFF can be trained in an
unsupervised learning fashion on diverse, free-form videos,
and applicable to addressing challenging tasks including video object segmentation,
pose tracking and long-range flow learning in terms of frame reconstruction.
We also visualize how each pixel evolves during solving these problems to gain better
interpretability of the model.

We evaluate our mgPFF model on the challenging video propagation tasks:
DAVIS2017~\cite{pont20172017} for video object segmentation and
long-range flow learning in terms of frame reconstruction,
and JHMDB dataset~\cite{jhuang2013towards} for human pose tracking.

\noindent{\bf Compared methods}
include the simplistic identity mapping (always copying the first frame labels),
SIFT flow~\cite{liu2009beyond} which is an off-the-shelf toolbox for dense
correspondence alignment,
learning-based optical flow (FlowNet2)~\cite{ilg2017flownet}  which is trained on large-scale
synthetic data,
DeepCluster~\cite{caron2018deep} which is unsupervised trained for
clustering on ImageNet~\cite{imagenet_cvpr09},
ColorPointer~\cite{vondrick2018tracking} which learns video colorization and shows effective in
object tracking,
and CycleTime~\cite{xiaolong2019learning} which exploits the cycle consistence along time
and is trained for patch reconstruction with mid-level feature activations.

\subsection{Experimental Setup}
\noindent{\bf Training.}
We train our mgPFF model from scratch over a combined
datasets consisting of the whole Sintel Movie~\cite{LevyR10Sintel},
training set of DAVIS2017~\cite{pont20172017},
and training set of JHMDB (split1)~\cite{jhuang2013towards}.
It is worth noting that our whole training set
contains only $\sim$6$\times$$10^{4}$ frames,
whereas our compared methods train over orders magnitude larger dataset.
For example,
ColorPointer~\cite{vondrick2018tracking} is trained over 300K videos
($\sim$9$\times$$10^{7}$ frames)
from Kinetics dataset~\cite{kay2017kinetics},
and CycleTime~\cite{xiaolong2019learning} is trained over 114K videos
(344-hour recording, $\sim$3.7$\times$$10^{7}$ frames)
from VLOG dataset~\cite{fouhey2018lifestyle}.
Moreover, most interestingly,
in training our mgPFF on the Sintel movie,
we find mgPFF automatically learns to detect the video
shot/transition~\cite{boreczky1996comparison, gargi2000performance} purely
based on the reconstruction errors between input frames (see Fig.~\ref{fig:shotDet}).

We use ADAM optimization method during training~\cite{kingma2014adam},
with initial learning 0.0005 and coefficients 0.9 and 0.999
for computing running averages of gradient and its square.
We randomly initialize the weights and train from scratch
over free form videos.
We train our model using PyTorch~\cite{paszke2017automatic}
on a single NVIDIA TITAN X GPU,
and terminate after 500K iteration updates.\footnote{
The code and models can be found in
\url{https://github.com/aimerykong/predictive-filter-flow}}
During training,
we randomly sample frame pairs (resized to 256$\times$256-pixel resolution) within
$N$=5 consecutive frames.
We also augment the training set by randomly flipping and rotating the frame pairs.
After training the model on the combined dataset,
we train specifically over the training set (without annotation) of DAVIS2017 and JHMDB
respectively
for video object segmentation and human pose tracking.

\noindent{\bf Inference.}
We essentially propagate the given mask/pose at the first frame along the time.
We also set the temporal window size $K$,
meaning we warp towards the target frame using previous $K$ frames.
We test different temporal window size for video segmentation and tracking
and find $K$=3 works the best.
Specifically,
for video object segmentation on DAVIS2017,
we threshold with 0.8 the propagated mask at each tracking update,
since pixels on the foreground (within the mask) may diffuse to background,
and filter flow gives probabilities around the mask boundary.
For human pose tracking,
we dilate the joints for propagation, and vote for the tracked joint after propagation
as the track. This gives stable tracking though sometimes the track may stay at the
background especially when the background is similar to the
foreground (3rd video in Fig.~\ref{fig:JHMDB_show}).
We note that there are other methods using low-level cues for higher-level tasks,
e.g., using boundary for semantic segmentation~\cite{arbelaez2011contour,
maninis2018convolutional}.

\subsection{Unsupervised Learning for Video Segmentation}

{
\setlength{\tabcolsep}{0.2em} 
\begin{table}
     \footnotesize
     \centering
\caption{\small \textbf{Tracking Segmentation} on the DAVIS2017 validation set.
Methods marked with \textcolor{red}{$^{1st}$} additionally use the first frame and
its mask (provided) for tracking in the rest of the video.
The \textcolor{blue}{number} in bracket is the estimated number of frames
used for training the corresponding method.
}
\vspace{-3mm}
   \begin{tabular}{l l | l | c c | c c }
	\hline
         &\multirow{2}{*}{Method} & \multirow{2}{*}{Supervision}
	& \multicolumn{2}{c|}{$\mathcal{J}$ (segments)} &  \multicolumn{2}{c}{$\mathcal{F}$ (boundaries)} \\
	\cline{4-7}
        & & & mean$\uparrow$ & recall$\uparrow$ & mean$\uparrow$ & recall$\uparrow$  \\
        \hline
        & OSVOS \cite{caelles2017one} & {\scriptsize ImageNet, DAVIS} & 55.1 & 60.2 & 62.1 & 71.3  \\
        & MaskTrack \cite{perazzi2017learning} & {\scriptsize  ImageNet, DAVIS} & 51.2 & 59.7 & 57.3 & 65.5 \\
        & OSVOS-B \cite{caelles2017one} & ImageNet & 18.5 & 15.9  & 30.0 & 20.0  \\
        & MaskTrack-B \cite{perazzi2017learning} & ImageNet & 35.3 & 37.8 &  36.4 & 36.0 \\
        & OSVOS-M \cite{yang2018efficient} & ImageNet & 36.4 & 34.8 &  39.5 & 35.3 \\
        \hline
        & Identity & None & 22.1 & 15.9 &  23.6 & 11.7   \\
        & SIFTflow~\cite{liu2009beyond} & None & 13.0 &  7.9 & 15.1 & 5.5   \\
        & SIFTflow\textcolor{red}{$^{1st}$}~\cite{liu2009beyond}
                        & None & 33.0 &  -- & 35.0 & --   \\
        & FlowNet2~\cite{ilg2017flownet} & Synthetic&  16.7  &  9.5  &  19.7  &  7.6 \\
        & FlowNet2\textcolor{red}{$^{1st}$}~\cite{ilg2017flownet} & Synthetic
                                         &  26.7  &  --  &  25.2  &  -- \\
        & DeepCluster\textcolor{red}{$^{1st}$}~\cite{caron2018deep} & Self \textcolor{blue}{(1.3$\times$$10^6$)}
						& 37.5  &  -- & 33.2  &  -- \\
        & ColorPointer~\cite{vondrick2018tracking} & Self \textcolor{blue}{(9.0$\times$$10^7$)} & 34.6  &  34.1 & 32.7  &  26.8 \\
        & CycleTime\textcolor{red}{$^{1st}$}~\cite{xiaolong2019learning} & Self \textcolor{blue}{(3.7$\times$$10^7$)}
                                                & 40.1  &  -- & 38.3  &  -- \\
        \hline
        & {\bf mgPFF} (1st only)  & \multirow{4}{*}{Self \textcolor{blue}{(6.0$\times$$10^4$)}} & 31.6  &  29.5
                        & 36.2 & 30.8  \\
        & {\bf mgPFF}  ($K$=1) &  & 38.9  &  38.5   & 41.1 & 38.6  \\
        & {\bf mgPFF}\textcolor{red}{$^{1st}$} ($K$=1)
                        &  & {41.9} & {41.4} & {45.2} & {43.9}   \\
        & {\bf mgPFF}\textcolor{red}{$^{1st}$} ($K$=3)
                        &  & {\bf 42.2} & {\bf 41.8} & {\bf 46.9} & {\bf44.4}   \\
        \hline
     \end{tabular}
\vspace{-3mm}
     \label{tab:davis2017}
 \end{table}
}

\begin{figure*}[t]
    \centering
    \begin{minipage}{0.99\textwidth}
        \centering
        \includegraphics[width=1\linewidth]{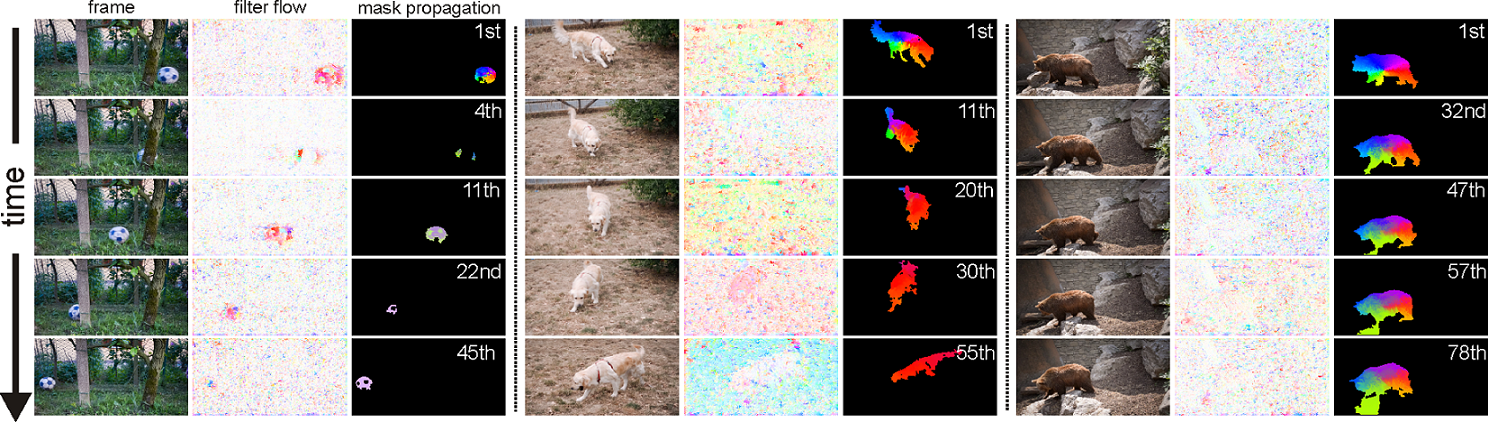}
        \captionsetup{width=0.999\textwidth}
    \end{minipage}
    \vspace{-4mm}
    \caption{\small Visualization of unsupervised learning for video segmentation on video from DAVIS2017:
    \emph{soccerball}, \emph{dog} and \emph{bear}.
    We show the tracking results with temporal window size $K$=3 for \emph{soccerball} (otherwise it loses track
    due to heavy occlusion)  and $K$=1 for others.
    Note that in \emph{soccerball}, there are heavy occlusions but our mgPFF model can still
    track the ball.
    In \emph{dog}, we can see how each pixel moves along with the dog:
    when the dog turns from right side to left side, the colors from the neck are propagated for tracking.
    This demonstrates how mgPFF tracks each pixel in the physical manifold flavor.
    In \emph{bear}, the disocclusion shadow arises from the bottom border of the image, connecting
    with the bear, then mgPFF propagates the bear leg to the shadow.
    }
    \label{fig:DAVIS_show}
    \vspace{-1mm}
\end{figure*}

We analyze our model on video segmentation over the DAVIS 2017 validation
set~\cite{pont20172017},
where the initial segmentation mask is given and the task is to predict the segmentation in the
rest of the video.
This is a very challenging task as the videos contain multiple objects that undergo
significant occlusion, deformation and scale change with clutter background,
as shown in Fig.~\ref{fig:DAVIS_show}.
We use the provided code and report two metrics that score segment overlap and boundary accuracy.
The Jacaard index $\mathcal{J}$ is defined as the intersection-over-union of the estimated
segmentation and the ground-truth mask, measuring how well the pixels of two masks
match~\cite{everingham2010pascal}. The $\mathcal{J}$ recall measures the fraction of sequences
with IoU$>$0.5.
The F-measure denoted by $\mathcal{F}$ considers both contour-based precision and recall that
measure the accuracy of the segment contours~\cite{martin2004learning}.

We compare our mgPFF with other unsupervised methods as well as some supervised
ones~\cite{yang2018efficient, caelles2017one}  in
Table~\ref{tab:davis2017}.
The first two supervised methods are trained explicitly
using the annotated masks along
with training video frames.
As in literature there are methods always using the given mask at the first frame to aid
tracking,
we also follow this practice with mgPFF to report the performance.
But before doing so,
we ablate how much gain we can get from using only the given mask for the tracking.
To this end,
we setup the mgPFF by always propagating the given mask for tracking,
as noted by mgPFF (1st only) in Table~\ref{tab:davis2017}.
Surprisingly,
this simple setup works very well, even better than flow based methods, such as
SIFTflow\textcolor{red}{$^{1st}$} and FlowNet2\textcolor{red}{$^{1st}$},
both of which not only use the first frame but also the previous $N$=4
frames for tracking.
This suggests the mgPFF is able to capture long-range flow even though we did not train our
model with frames across large intervals.
We explicitly study this long-range flow in Section~\ref{ssec:long-range-flow} quantitatively.

When we perform tracking with the \emph{only one} previous propagated mask ($K$=1),
our mgPFF outperforms all the other unsupervised methods,
except CycleTime (on $\cal J$ measure only),
which is explicitly trained at patch level thus captures better object segment.
When additionally using the mask given at the first frame for tracking in
subsequent frames,
mgPFF\textcolor{red}{$^{1st}$}($K$=1) outperforms all other unsupervised methods by a notable margin,
and our mgPFF\textcolor{red}{$^{1st}$}($K$=3) achieves the best performance.
In particuar, in terms of the boundary measure,
we can see mgPFF performs significantly better than the
other unsupervised methods. This demonstrates the benefit of propagating masks
with fine-grained pixel-level flows instead of flows learned at patch level through mid-level feature
activations~\cite{vondrick2018tracking, xiaolong2019learning}.

Overall,
we note that our mgPFF even outperforms several supervised methods,
but only worse than the first two
supervised models in Table~\ref{tab:davis2017} which are
explicitly trained with DAVIS pixel-level annotations at all training frames.
Moreover,
it is worth noting that our mgPFF model is trained over two orders magnitude less
data than other unsupervised methods, e.g., DeepCluster, ColorPointer and CycleTime.
This demonstrates the benefit brought by the low-vision nature of mgPFF that it does
not demand very large-scale training data.

In Fig.~\ref{fig:DAVIS_show},
we visualize the tracking results ($N$=3)
and the predicted filter flow (from previous one frame only).
Specifically,
we transform the filter flow into the flow vector (Eq.~\ref{eq:ff2of}) and
treat this as
optical flow for visualization.
As mgPFF performs at pixel level,
we are able to visualize the tracking through more fine-grained details.
We paint on the mask with the color chart from optical flow,
and visualize to see how the pixels evolve over time.
Interestingly,
from this visualization,
we can see how tracking is accomplished in front of heavy occlusion,
big deformation and similar background situation (see descriptions
under Fig.~\ref{fig:DAVIS_show}).


\subsection{Unsupervised Learning for Pose Tracking}

{
\setlength{\tabcolsep}{0.5em} 
\begin{table}
    \small
    \centering
    \caption{\textbf{Human Pose Tracking} on JHMDB dataset.
    Methods marked with \textcolor{red}{$^{1st}$} additionally use the first frame with its mask
    for propagating on the rest frames.
    ``mgPFF+ft'' means that we fine-tune mgPFF model particularly on the videos from this dataset
    in an unsupervised way (no annotations used).
    }
    \vspace{-1mm}
    \begin{tabular}{l | c  c  c  c c}
    \hline
    Method / PCK$\uparrow$  & @0.1 & @0.2 & @0.3 & @0.4 & @0.5 \\
    \hline
    fully-supervised~\cite{song2017thin} & 68.7 & 92.1 \\
    \hline
    Identity & 43.1 & 64.5 & 76.0 & 83.5 & 88.5 \\
    SIFTflow\textcolor{red}{$^{1st}$}~\cite{liu2009beyond} & 49.0 & 68.6 & -- & -- & -- \\
    FlowNet2~\cite{ilg2017flownet} & 45.2 & 62.9 & 73.5 & 80.6 & 85.5 \\
    DeepCluster\textcolor{red}{$^{1st}$}~\cite{caron2018deep} & 43.2 & 66.9 \\
    ColorPointer~\cite{vondrick2018tracking} & 45.2 & 69.6 & 80.8 & 87.5 & 91.4 \\
    CycleTime\textcolor{red}{$^{1st}$}~\cite{xiaolong2019learning}
                & 57.3 & 78.1 & -- & -- & -- \\
    \hline
    {\bf mgPFF} & 49.3 & 72.8 & 82.4 & 88.6 & 91.9 \\
    {\bf mgPFF}\textcolor{red}{$^{1st}$} & 55.6 & 77.1 & 85.2 & 89.6 & 92.1 \\
    {\bf mgPFF}+ft & 52.7 & 75.1 & 84.0 & 89.5 & 92.3 \\
    {\bf mgPFF}+ft\textcolor{red}{$^{1st}$} & {\bf 58.4} & {\bf 78.1} & {\bf 85.9} & {\bf 89.8} & {\bf 92.4} \\
    \hline
    \end{tabular}
    \vspace{-2mm}
    \label{tab:jhmdb}
\end{table}
}

\begin{figure*}[t]
    \centering
    \begin{minipage}{0.99\textwidth}
        \centering
        \includegraphics[width=1\linewidth]{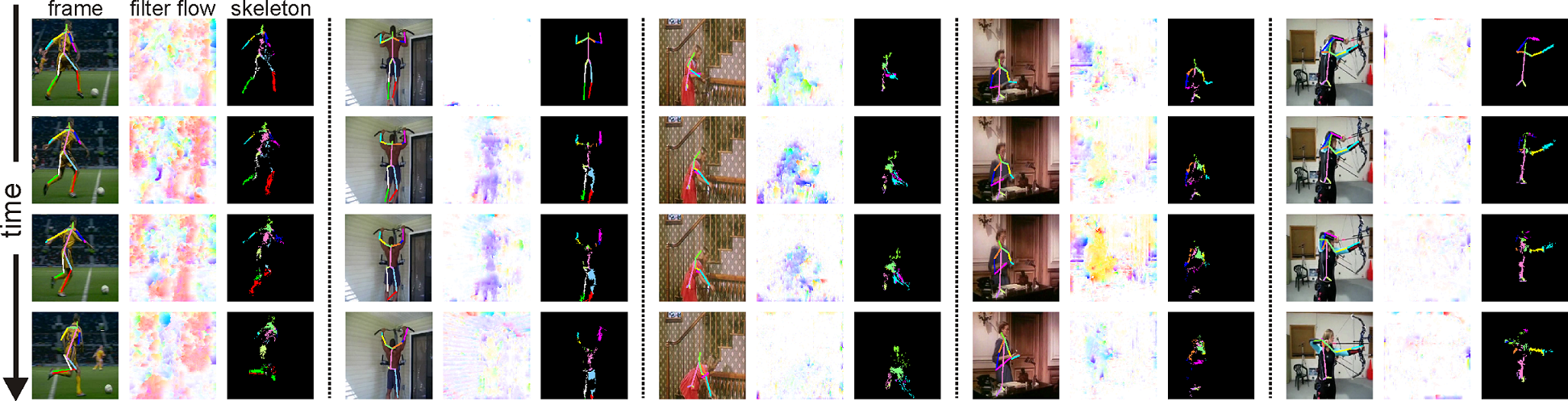}
        \captionsetup{width=0.999\textwidth}
    \end{minipage}
    \vspace{-2mm}
    \caption{\small {\bf Human Pose Tracking}
    on JHMDB dataset.
    We show results by propagating only the previous mask ($K$=1),
    and overlay the tracked joints on the RGB frames.
    Besides, with the predicted filter flow,
    we also propagate the colorful skeleton to visualize how pixels on
    the skeleton evolve over time.
    In last row,
    we pick the results around the end of each video to show how mgPFF
    fails in tracking,
    mainly due to heavy occlusion (knees in the 1st video),
    joints moving outside the image (ankle in the 2nd video),
    similar background (hair color in the 3rd video),
    and motion blur (elbow in the 4th video).
    (Best viewed in color and zoom-in.)
    }
    \label{fig:JHMDB_show}
    \vspace{-3mm}
\end{figure*}

We validate our mgPFF for human pose tracking on the JHMDB
dataset~\cite{jhuang2013towards}.
During testing,
we are given an initial frame labeled with 15 human joints and the task
is to predict the joints in the subsequent frames.
To this end,
we stack the 15 maps for the 15 joints as a 3D array,
and propagate the array using the predicted filter flow.
To report performance,
we use the Probability of Correct Keypoint (PCK@$\tau$) from~\cite{yang2013articulated},
which measures the portion of predicted points that are within a radius to their ground-truth,
where the radius is $\tau$ times the size of the human pose bounding box.

In Table~\ref{tab:jhmdb} we list the performance of different unsupervised learning methods,
and report two setups of the mgPFF on the validation set (split1):
1) with the model trained on the combined dataset,
and 2) with the model further fine-tuned
on JHMDB in an unsupervised
way~\cite{ranjan2017optical, wang2018occlusion, hirsch2010efficient}.
Similar to video segmentation,
without using the provided joints at the first frame for all subsequent tracking,
our mgPFF outperforms all other methods except CycleTime which always uses the first frame
(with the provided keypoints) for pose tracking.
By fine-tuning our model on the videos of this dataset (without using the joint annotations),
we obtain further improvement;
but the improvement is less than additionally using the first frame for tracking.
We conjecture the reason is that by using the provided mask at the first frame,
mgPFF is able to warp all the available joints toward current frame;
otherwise it may lose track once the joints move outside the image (see 2nd video in
Fig.~\ref{fig:JHMDB_show}).
It is worth noting that
mgPFF as well as the learning based optical flow method performs fast in propagating
the joints for tracking,
whereas DeepCluster, ColorPointer and CycleTime require computing affinity matrix over
all pixels from previous $K$ frames~\cite{vondrick2018tracking, xiaolong2019learning}.
Moreover, although it seems unfair to compare our mgPFF with unsupervised fine-tuning on
the same JHMDB dataset, we note that
ColorPointer and CycleTime are trained on much larger dataset consisting mainly of
human actions/activities.

In Fig.~\ref{fig:JHMDB_show},
we visualize the pose tracking results as well as the filter flow and how each pixel along
the skeleton evolves over time.
We plot in last row the frames on which our mgPFF starts to fail in tracking.
The failure cases are largely due to challenging situations,
like heavy occlusion (1st video),
joint moving outside the image,
similar background (3rd video) and big motion blur (4th video).

\subsection{Long-Range Flow for Frame Reconstruction}
\label{ssec:long-range-flow}

\begin{figure}[t]
\small
    \centering
    \begin{minipage}{0.45\textwidth}
        \centering
        \includegraphics[width=1\linewidth]{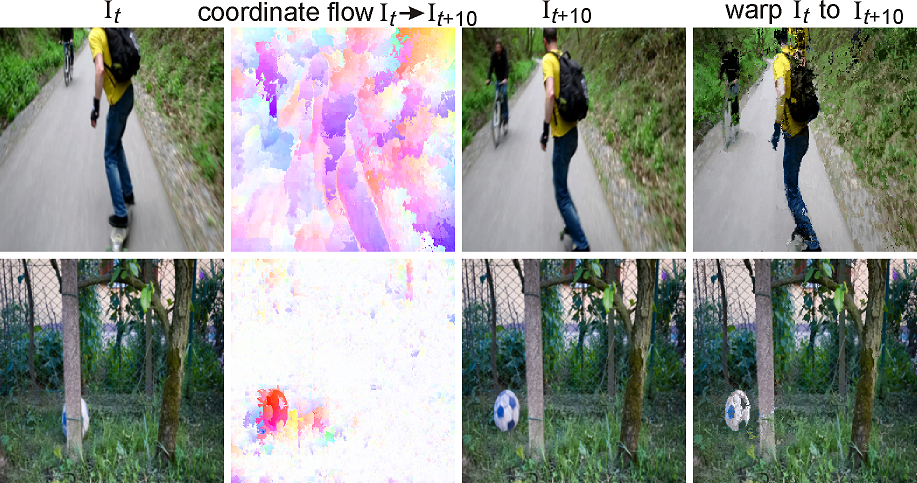}
        \captionsetup{width=0.999\textwidth}
    \end{minipage}
    \vspace{-1mm}
    \caption{\small {\bf Long-range flow for frame reconstruction}
    (rightmost column)
    by warping $\I_t$ (1st column) with the coordinate flow (2nd column)
    which is transformed from the predicted multigrid filter flow.
    The target frames $\I_{t+10}$ are shown in the 3rd column.
    }
    \label{fig:frameReconstruction}
    \vspace{-2mm}
\end{figure}

{
\setlength{\tabcolsep}{0.5em} 
\begin{table}[t]
\small
\centering
\caption{{\bf Long-range flow for frame Reconstruction}:
We compute the long-range flow on two frames and warp the 1st one with the flow.
We compare the warped frame with the 2nd frame measured by pixel-level L1 distance.
The gaps are 5 and 10, respectively.}
\vspace{-2mm}
\begin{tabular}{ l | c  | c    }
\hline
   method/error$\downarrow$ & 5-Frame & 10-Frame\\
   \hline
   Identity                         & 82.0 & 97.7 \\
   Optical Flow (FlowNet2)~\cite{ilg2017flownet}   & 62.4 & 90.3 \\
   CycleTime~\cite{xiaolong2019learning} & 60.4 & 76.4 \\
   {\bf mgPFF}                            & \bf{7.32} & {\bf 8.83} \\
   \hline
\end{tabular}
\vspace{-2mm}
\label{tab:frameReconstruction}
\end{table}
}

We highlight our mgPFF is particularly good at learning long-range flow
for reconstructing frames.
To validate this,
specifically,
given two frames $\I_t$ and $\I_{t+m}$ distant in time in a video,
we predict the filter flow between them,
and then transform the filter flow into coordinate flow
according to Eq.~\ref{eq:ff2of} to indicate
where to copy pixels from $\I_t$.
With the coordinate flow,
we warp frame $\I_t$ to generate a new frame  $\hat\I_{t+m}$.
We compare the pixel-level L1 distance between $\I_{t}$ and $\hat\I_{t+m}$ in
original uint8 RGB space ([0,255] scale).

We perform this experiment on DAVIS2017 validation set,
and report the performance in Table~\ref{tab:frameReconstruction},
in which we set the time gap as $m$=5 or $m$=10, meaning the two frames
are $m$ frames apart from each other.
In both frame gaps,
our mgPFF significantly outperforms the compared methods,
demonstrating the powerfulness of mgPFF in modeling pixel level movement,
even though our model is trained over frame pairs within 5-frame interval without seeing
any frames far away from 5 frames.
In Fig.~\ref{fig:frameReconstruction},
we clearly see that mgPFF performs quite well visually on long-range flow learning
for frame reconstruction.

\section{Conclusion}
We propose a simple, compact framework for unsupervised learning on free-form videos,
named multigrid Predictive Filter Flow (mgPFF).
Through experiments, we show mgPFF outperforms other state-of-the-art methods notably
in video object segmentation and human pose tracking with the unsupervised learning setup;
it also exhibits great power in long-range flow learning in terms of frame
reconstruction.
In this sense, it is reminiscent of a variety of other flow-based tasks, such as video
compression~\cite{rippel2018learned}, frame interpolation~\cite{niklaus2017video},
unsupervised optical flow learning~\cite{yang2018every, janai2018unsupervised},
etc., which are all candidates for future extensions.
Moreover, based on the filter flow output which is fast in computation,
it is also interesting to use it for action classification where
the flow stream consistently improves performance~\cite{simonyan2014two}
but optical flow estimation is slow.
Finally the pixel embedding features~\cite{kong2018recurrent, wei2018learning}
could also be used as video frame representation for
action classification~\cite{misra2016shuffle, fernando2017self}.
We expect further improvement on these tasks by taking as mgPFF as initial
proposal generation with followup mechanisms for fine video
segmentation~\cite{xie2018object, vondrick2018tracking, xiaolong2019learning, pathak2017learning}.

\section*{Acknowledgement}
{ \small
This project is supported by NSF grants IIS-1813785, IIS-1618806, IIS-1253538
and a hardware donation from NVIDIA.
Shu Kong personally thanks Teng Liu and Etthew Kong who initiated this research,
and the academic uncle Alexei A. Efros for the encouragement and discussion.
}

{\small
\bibliographystyle{ieee}
\bibliography{egbib}

\begin{thebibliography}{100}\itemsep=-1pt

\bibitem{Abdulla2018hiddenLayer}
W.~Abdulla and P.~Ferriere.
\newblock Neural network graphs and training metrics for pytorch and
  tensorflow.
\newblock 2018.

\bibitem{abu2016youtube}
S.~Abu-El-Haija, N.~Kothari, J.~Lee, P.~Natsev, G.~Toderici, B.~Varadarajan,
  and S.~Vijayanarasimhan.
\newblock Youtube-8m: A large-scale video classification benchmark.
\newblock {\em arXiv preprint arXiv:1609.08675}, 2016.

\bibitem{agrawal2016learning}
P.~Agrawal, A.~V. Nair, P.~Abbeel, J.~Malik, and S.~Levine.
\newblock Learning to poke by poking: Experiential learning of intuitive
  physics.
\newblock In {\em Advances in Neural Information Processing Systems}, pages
  5074--5082, 2016.

\bibitem{arbelaez2011contour}
P.~Arbelaez, M.~Maire, C.~Fowlkes, and J.~Malik.
\newblock Contour detection and hierarchical image segmentation.
\newblock {\em IEEE transactions on pattern analysis and machine intelligence},
  33(5):898--916, 2011.

\bibitem{boreczky1996comparison}
J.~S. Boreczky and L.~A. Rowe.
\newblock Comparison of video shot boundary detection techniques.
\newblock {\em Journal of Electronic Imaging}, 5(2):122--129, 1996.

\bibitem{BruhnW05}
A.~Bruhn and J.~Weickert.
\newblock Towards ultimate motion estimation: Combining highest accuracy with
  real-time performance.
\newblock In {\em 10th {IEEE} International Conference on Computer Vision
  {(ICCV} 2005), 17-20 October 2005, Beijing, China}, pages 749--755, 2005.

\bibitem{caba2015activitynet}
F.~Caba~Heilbron, V.~Escorcia, B.~Ghanem, and J.~Carlos~Niebles.
\newblock Activitynet: A large-scale video benchmark for human activity
  understanding.
\newblock In {\em Proceedings of the IEEE Conference on Computer Vision and
  Pattern Recognition}, pages 961--970, 2015.

\bibitem{caelles2017one}
S.~Caelles, K.-K. Maninis, J.~Pont-Tuset, L.~Leal-Taix{\'e}, D.~Cremers, and
  L.~Van~Gool.
\newblock One-shot video object segmentation.
\newblock In {\em Proceedings of the IEEE conference on computer vision and
  pattern recognition}, pages 221--230, 2017.

\bibitem{caron2018deep}
M.~Caron, P.~Bojanowski, A.~Joulin, and M.~Douze.
\newblock Deep clustering for unsupervised learning of visual features.
\newblock In {\em Proceedings of the European Conference on Computer Vision
  (ECCV)}, pages 132--149, 2018.

\bibitem{cordts2016cityscapes}
M.~Cordts, M.~Omran, S.~Ramos, T.~Rehfeld, M.~Enzweiler, R.~Benenson,
  U.~Franke, S.~Roth, and B.~Schiele.
\newblock The cityscapes dataset for semantic urban scene understanding.
\newblock In {\em Proceedings of the IEEE conference on computer vision and
  pattern recognition}, pages 3213--3223, 2016.

\bibitem{imagenet_cvpr09}
J.~Deng, W.~Dong, R.~Socher, L.-J. Li, K.~Li, and L.~Fei-Fei.
\newblock {ImageNet: A Large-Scale Hierarchical Image Database}.
\newblock In {\em CVPR09}, 2009.

\bibitem{doersch2015unsupervised}
C.~Doersch, A.~Gupta, and A.~A. Efros.
\newblock Unsupervised visual representation learning by context prediction.
\newblock In {\em Proceedings of the IEEE International Conference on Computer
  Vision}, pages 1422--1430, 2015.

\bibitem{doersch2017multi}
C.~Doersch and A.~Zisserman.
\newblock Multi-task self-supervised visual learning.
\newblock In {\em Proceedings of the IEEE International Conference on Computer
  Vision}, pages 2051--2060, 2017.

\bibitem{dosovitskiy2015flownet}
A.~Dosovitskiy, P.~Fischer, E.~Ilg, P.~Hausser, C.~Hazirbas, V.~Golkov, P.~Van
  Der~Smagt, D.~Cremers, and T.~Brox.
\newblock Flownet: Learning optical flow with convolutional networks.
\newblock In {\em Proceedings of the IEEE International Conference on Computer
  Vision}, pages 2758--2766, 2015.

\bibitem{everingham2010pascal}
M.~Everingham, L.~Van~Gool, C.~K. Williams, J.~Winn, and A.~Zisserman.
\newblock The pascal visual object classes (voc) challenge.
\newblock {\em International journal of computer vision}, 88(2):303--338, 2010.

\bibitem{fernando2017self}
B.~Fernando, H.~Bilen, E.~Gavves, and S.~Gould.
\newblock Self-supervised video representation learning with odd-one-out
  networks.
\newblock In {\em Proceedings of the IEEE conference on computer vision and
  pattern recognition}, pages 3636--3645, 2017.

\bibitem{fleet2006optical}
D.~Fleet and Y.~Weiss.
\newblock Optical flow estimation.
\newblock In {\em Handbook of mathematical models in computer vision}, pages
  237--257. Springer, 2006.

\bibitem{fouhey2018lifestyle}
D.~F. Fouhey, W.-c. Kuo, A.~A. Efros, and J.~Malik.
\newblock From lifestyle vlogs to everyday interactions.
\newblock In {\em Proceedings of the IEEE Conference on Computer Vision and
  Pattern Recognition}, pages 4991--5000, 2018.

\bibitem{gargi2000performance}
U.~Gargi, R.~Kasturi, and S.~H. Strayer.
\newblock Performance characterization of video-shot-change detection methods.
\newblock {\em IEEE transactions on circuits and systems for video technology},
  10(1):1--13, 2000.

\bibitem{geiger2013vision}
A.~Geiger, P.~Lenz, C.~Stiller, and R.~Urtasun.
\newblock Vision meets robotics: The kitti dataset.
\newblock {\em The International Journal of Robotics Research},
  32(11):1231--1237, 2013.

\bibitem{godard2017unsupervised}
C.~Godard, O.~Mac~Aodha, and G.~J. Brostow.
\newblock Unsupervised monocular depth estimation with left-right consistency.
\newblock In {\em Proceedings of the IEEE Conference on Computer Vision and
  Pattern Recognition}, pages 270--279, 2017.

\bibitem{gold1962arrow}
T.~Gold.
\newblock The arrow of time.
\newblock {\em American Journal of Physics}, 30(6):403--410, 1962.

\bibitem{goren1975visual}
C.~C. Goren, M.~Sarty, and P.~Y. Wu.
\newblock Visual following and pattern discrimination of face-like stimuli by
  newborn infants.
\newblock {\em Pediatrics}, 56(4):544--549, 1975.

\bibitem{hackbusch2013multi}
W.~Hackbusch.
\newblock {\em Multi-grid methods and applications}, volume~4.
\newblock Springer Science \& Business Media, 2013.

\bibitem{he2016deep}
K.~He, X.~Zhang, S.~Ren, and J.~Sun.
\newblock Deep residual learning for image recognition.
\newblock In {\em Proceedings of the IEEE conference on computer vision and
  pattern recognition}, pages 770--778, 2016.

\bibitem{hirsch2011fast}
M.~Hirsch, C.~J. Schuler, S.~Harmeling, and B.~Sch{\"o}lkopf.
\newblock Fast removal of non-uniform camera shake.
\newblock In {\em 2011 International Conference on Computer Vision}, pages
  463--470. IEEE, 2011.

\bibitem{hirsch2010efficient}
M.~Hirsch, S.~Sra, B.~Sch{\"o}lkopf, and S.~Harmeling.
\newblock Efficient filter flow for space-variant multiframe blind
  deconvolution.
\newblock In {\em 2010 IEEE Computer Society Conference on Computer Vision and
  Pattern Recognition}, pages 607--614. IEEE, 2010.

\bibitem{hur2017mirrorflow}
J.~Hur and S.~Roth.
\newblock Mirrorflow: Exploiting symmetries in joint optical flow and occlusion
  estimation.
\newblock In {\em Proceedings of the IEEE International Conference on Computer
  Vision}, pages 312--321, 2017.

\bibitem{ilg2017flownet}
E.~Ilg, N.~Mayer, T.~Saikia, M.~Keuper, A.~Dosovitskiy, and T.~Brox.
\newblock Flownet 2.0: Evolution of optical flow estimation with deep networks.
\newblock In {\em Proceedings of the IEEE Conference on Computer Vision and
  Pattern Recognition}, pages 2462--2470, 2017.

\bibitem{ioffe2015batch}
S.~Ioffe and C.~Szegedy.
\newblock Batch normalization: Accelerating deep network training by reducing
  internal covariate shift.
\newblock {\em arXiv preprint arXiv:1502.03167}, 2015.

\bibitem{jaderberg2015spatial}
M.~Jaderberg, K.~Simonyan, A.~Zisserman, et~al.
\newblock Spatial transformer networks.
\newblock In {\em Advances in neural information processing systems}, pages
  2017--2025, 2015.

\bibitem{janai2018unsupervised}
J.~Janai, F.~Guney, A.~Ranjan, M.~Black, and A.~Geiger.
\newblock Unsupervised learning of multi-frame optical flow with occlusions.
\newblock In {\em Proceedings of the European Conference on Computer Vision
  (ECCV)}, pages 690--706, 2018.

\bibitem{jason2016back}
J.~Y. Jason, A.~W. Harley, and K.~G. Derpanis.
\newblock Back to basics: Unsupervised learning of optical flow via brightness
  constancy and motion smoothness.
\newblock In {\em European Conference on Computer Vision}, pages 3--10.
  Springer, 2016.

\bibitem{jayaraman2015learning}
D.~Jayaraman and K.~Grauman.
\newblock Learning image representations tied to ego-motion.
\newblock In {\em Proceedings of the IEEE International Conference on Computer
  Vision}, pages 1413--1421, 2015.

\bibitem{jhuang2013towards}
H.~Jhuang, J.~Gall, S.~Zuffi, C.~Schmid, and M.~J. Black.
\newblock Towards understanding action recognition.
\newblock In {\em Proceedings of the IEEE international conference on computer
  vision}, pages 3192--3199, 2013.

\bibitem{kay2017kinetics}
W.~Kay, J.~Carreira, K.~Simonyan, B.~Zhang, C.~Hillier, S.~Vijayanarasimhan,
  F.~Viola, T.~Green, T.~Back, P.~Natsev, M.~Suleyman, and A.~Zisserman.
\newblock The kinetics human action video dataset.
\newblock {\em arXiv preprint arXiv:1705.06950}, 2017.

\bibitem{perazzi2017learning}
A.~Khoreva, F.~Perazzi, R.~Benenson, B.~Schiele, and A.~Sorkine-Hornung.
\newblock Learning video object segmentation from static images.
\newblock {\em CoRR}, abs/1612.02646, 2016.

\bibitem{kingma2014adam}
D.~P. Kingma and J.~Ba.
\newblock Adam: A method for stochastic optimization.
\newblock {\em arXiv preprint arXiv:1412.6980}, 2014.

\bibitem{kong2018image}
S.~Kong and C.~Fowlkes.
\newblock Image reconstruction with predictive filter flow.
\newblock {\em arXiv preprint arXiv:1811.11482}, 2018.

\bibitem{kong2018recurrent}
S.~Kong and C.~Fowlkes.
\newblock Recurrent pixel embedding for instance grouping.
\newblock In {\em Proceedings of the IEEE Conference on Computer Vision and
  Pattern Recognition (CVPR)}, pages 9018--9028, 2018.

\bibitem{larsson2017colorization}
G.~Larsson, M.~Maire, and G.~Shakhnarovich.
\newblock Colorization as a proxy task for visual understanding.
\newblock In {\em Proceedings of the IEEE Conference on Computer Vision and
  Pattern Recognition}, pages 6874--6883, 2017.

\bibitem{lazebnik2004semi}
S.~Lazebnik, C.~Schmid, and J.~Ponce.
\newblock Semi-local affine parts for object recognition.
\newblock In {\em British Machine Vision Conference (BMVC'04)}, pages 779--788.
  The British Machine Vision Association (BMVA), 2004.

\bibitem{levin2009understanding}
A.~Levin, Y.~Weiss, F.~Durand, and W.~T. Freeman.
\newblock Understanding and evaluating blind deconvolution algorithms.
\newblock In {\em 2009 IEEE Conference on Computer Vision and Pattern
  Recognition}, pages 1964--1971. IEEE, 2009.

\bibitem{LevyR10Sintel}
C.~Levy and T.~Roosendaal.
\newblock Sintel.
\newblock In {\em {ACM} {SIGGRAPH} {ASIA} 2010 Computer Animation Festival,
  Seoul, Republic of Korea, December 15 - 18, 2010}, page 82:1, 2010.

\bibitem{li2013bayesian}
C.~Li, S.~Su, Y.~Matsushita, K.~Zhou, and S.~Lin.
\newblock Bayesian depth-from-defocus with shading constraints.
\newblock In {\em Proceedings of the IEEE Conference on Computer Vision and
  Pattern Recognition}, pages 217--224, 2013.

\bibitem{liu2009beyond}
C.~Liu et~al.
\newblock {\em Beyond pixels: exploring new representations and applications
  for motion analysis}.
\newblock PhD thesis, Massachusetts Institute of Technology, 2009.

\bibitem{maninis2018convolutional}
K.-K. Maninis, J.~Pont-Tuset, P.~Arbel{\'a}ez, and L.~Van~Gool.
\newblock Convolutional oriented boundaries: From image segmentation to
  high-level tasks.
\newblock {\em IEEE transactions on pattern analysis and machine intelligence},
  40(4):819--833, 2018.

\bibitem{martin2004learning}
D.~R. Martin, C.~C. Fowlkes, and J.~Malik.
\newblock Learning to detect natural image boundaries using local brightness,
  color, and texture cues.
\newblock {\em IEEE Transactions on Pattern Analysis \& Machine Intelligence},
  (5):530--549, 2004.

\bibitem{mei2013segment}
X.~Mei, X.~Sun, W.~Dong, H.~Wang, and X.~Zhang.
\newblock Segment-tree based cost aggregation for stereo matching.
\newblock In {\em Proceedings of the IEEE Conference on Computer Vision and
  Pattern Recognition}, pages 313--320, 2013.

\bibitem{meister2018unflow}
S.~Meister, J.~Hur, and S.~Roth.
\newblock Unflow: Unsupervised learning of optical flow with a bidirectional
  census loss.
\newblock In {\em Thirty-Second AAAI Conference on Artificial Intelligence},
  2018.

\bibitem{menze2015discrete}
M.~Menze, C.~Heipke, and A.~Geiger.
\newblock Discrete optimization for optical flow.
\newblock In {\em German Conference on Pattern Recognition}, pages 16--28.
  Springer, 2015.

\bibitem{mildenhall2018burst}
B.~Mildenhall, J.~T. Barron, J.~Chen, D.~Sharlet, R.~Ng, and R.~Carroll.
\newblock Burst denoising with kernel prediction networks.
\newblock In {\em Proceedings of the IEEE Conference on Computer Vision and
  Pattern Recognition}, pages 2502--2510, 2018.

\bibitem{misra2016shuffle}
I.~Misra, C.~L. Zitnick, and M.~Hebert.
\newblock Shuffle and learn: unsupervised learning using temporal order
  verification.
\newblock In {\em European Conference on Computer Vision}, pages 527--544.
  Springer, 2016.

\bibitem{moore1978visual}
M.~K. Moore, R.~Borton, and B.~L. Darby.
\newblock Visual tracking in young infants: Evidence for object identity or
  object permanence?
\newblock {\em Journal of Experimental Child Psychology}, 25(2):183--198, 1978.

\bibitem{muller1978visual}
A.~A. Muller and R.~N. Aslin.
\newblock Visual tracking as an index of the object concept.
\newblock {\em Infant Behavior and Development}, 1:309--319, 1978.

\bibitem{nair2010rectified}
V.~Nair and G.~E. Hinton.
\newblock Rectified linear units improve restricted boltzmann machines.
\newblock In {\em Proceedings of the 27th international conference on machine
  learning (ICML-10)}, pages 807--814, 2010.

\bibitem{newcombe2015dynamicfusion}
R.~A. Newcombe, D.~Fox, and S.~M. Seitz.
\newblock Dynamicfusion: Reconstruction and tracking of non-rigid scenes in
  real-time.
\newblock In {\em Proceedings of the IEEE conference on computer vision and
  pattern recognition}, pages 343--352, 2015.

\bibitem{newell2017associative}
A.~Newell, Z.~Huang, and J.~Deng.
\newblock Associative embedding: End-to-end learning for joint detection and
  grouping.
\newblock In {\em Advances in Neural Information Processing Systems}, pages
  2277--2287, 2017.

\bibitem{niklaus2017videoAdaptiveConv}
S.~Niklaus, L.~Mai, and F.~Liu.
\newblock Video frame interpolation via adaptive convolution.
\newblock In {\em Proceedings of the IEEE Conference on Computer Vision and
  Pattern Recognition}, pages 670--679, 2017.

\bibitem{niklaus2017video}
S.~Niklaus, L.~Mai, and F.~Liu.
\newblock Video frame interpolation via adaptive separable convolution.
\newblock In {\em Proceedings of the IEEE International Conference on Computer
  Vision}, pages 261--270, 2017.

\bibitem{noroozi2016unsupervised}
M.~Noroozi and P.~Favaro.
\newblock Unsupervised learning of visual representations by solving jigsaw
  puzzles.
\newblock In {\em European Conference on Computer Vision}, pages 69--84.
  Springer, 2016.

\bibitem{owens2016ambient}
A.~Owens, J.~Wu, J.~H. McDermott, W.~T. Freeman, and A.~Torralba.
\newblock Ambient sound provides supervision for visual learning.
\newblock In {\em European conference on computer vision}, pages 801--816.
  Springer, 2016.

\bibitem{paszke2017automatic}
A.~Paszke, S.~Gross, S.~Chintala, G.~Chanan, E.~Yang, Z.~DeVito, Z.~Lin,
  A.~Desmaison, L.~Antiga, and A.~Lerer.
\newblock Automatic differentiation in pytorch.
\newblock 2017.

\bibitem{pathak2017learning}
D.~Pathak, R.~Girshick, P.~Doll{\'a}r, T.~Darrell, and B.~Hariharan.
\newblock Learning features by watching objects move.
\newblock In {\em Proceedings of the IEEE Conference on Computer Vision and
  Pattern Recognition}, pages 2701--2710, 2017.

\bibitem{pathak2016context}
D.~Pathak, P.~Krahenbuhl, J.~Donahue, T.~Darrell, and A.~A. Efros.
\newblock Context encoders: Feature learning by inpainting.
\newblock In {\em Proceedings of the IEEE conference on computer vision and
  pattern recognition}, pages 2536--2544, 2016.

\bibitem{perrone2016clearer}
D.~Perrone and P.~Favaro.
\newblock A clearer picture of total variation blind deconvolution.
\newblock {\em IEEE transactions on pattern analysis and machine intelligence},
  38(6):1041--1055, 2016.

\bibitem{pinto2016curious}
L.~Pinto, D.~Gandhi, Y.~Han, Y.-L. Park, and A.~Gupta.
\newblock The curious robot: Learning visual representations via physical
  interactions.
\newblock In {\em European Conference on Computer Vision}, pages 3--18.
  Springer, 2016.

\bibitem{pohlen2017full}
T.~Pohlen, A.~Hermans, M.~Mathias, and B.~Leibe.
\newblock Full-resolution residual networks for semantic segmentation in street
  scenes.
\newblock In {\em Proceedings of the IEEE Conference on Computer Vision and
  Pattern Recognition}, pages 4151--4160, 2017.

\bibitem{pont20172017}
J.~Pont-Tuset, F.~Perazzi, S.~Caelles, P.~Arbel{\'a}ez, A.~Sorkine-Hornung, and
  L.~Van~Gool.
\newblock The 2017 davis challenge on video object segmentation.
\newblock {\em arXiv preprint arXiv:1704.00675}, 2017.

\bibitem{popper1956arrow}
K.~R. Popper.
\newblock The arrow of time.
\newblock {\em Nature}, 177(4507):538, 1956.

\bibitem{ranjan2017optical}
A.~Ranjan and M.~J. Black.
\newblock Optical flow estimation using a spatial pyramid network.
\newblock In {\em Proceedings of the IEEE Conference on Computer Vision and
  Pattern Recognition}, pages 4161--4170, 2017.

\bibitem{ravi2017filter}
S.~N. Ravi, Y.~Xiong, L.~Mukherjee, and V.~Singh.
\newblock Filter flow made practical: Massively parallel and lock-free.
\newblock In {\em Proceedings of the IEEE Conference on Computer Vision and
  Pattern Recognition}, pages 3549--3558, 2017.

\bibitem{ren2017unsupervised}
Z.~Ren, J.~Yan, B.~Ni, B.~Liu, X.~Yang, and H.~Zha.
\newblock Unsupervised deep learning for optical flow estimation.
\newblock In {\em Thirty-First AAAI Conference on Artificial Intelligence},
  2017.

\bibitem{revaud2015epicflow}
J.~Revaud, P.~Weinzaepfel, Z.~Harchaoui, and C.~Schmid.
\newblock Epicflow: Edge-preserving interpolation of correspondences for
  optical flow.
\newblock In {\em Proceedings of the IEEE conference on computer vision and
  pattern recognition}, pages 1164--1172, 2015.

\bibitem{rhodin2018unsupervised}
H.~Rhodin, M.~Salzmann, and P.~Fua.
\newblock Unsupervised geometry-aware representation for 3d human pose
  estimation.
\newblock In {\em Proceedings of the European Conference on Computer Vision
  (ECCV)}, pages 750--767, 2018.

\bibitem{rippel2018learned}
O.~Rippel, S.~Nair, C.~Lew, S.~Branson, A.~G. Anderson, and L.~Bourdev.
\newblock Learned video compression.
\newblock {\em arXiv preprint arXiv:1811.06981}, 2018.

\bibitem{ronneberger2015u}
O.~Ronneberger, P.~Fischer, and T.~Brox.
\newblock U-net: Convolutional networks for biomedical image segmentation.
\newblock In {\em International Conference on Medical image computing and
  computer-assisted intervention}, pages 234--241. Springer, 2015.

\bibitem{scharstein2002taxonomy}
D.~Scharstein and R.~Szeliski.
\newblock A taxonomy and evaluation of dense two-frame stereo correspondence
  algorithms.
\newblock {\em International journal of computer vision}, 47(1-3):7--42, 2002.

\bibitem{seitz2009filter}
S.~M. Seitz and S.~Baker.
\newblock Filter flow.
\newblock In {\em Proceedings of the IEEE Conference on Computer Vision and
  Pattern Recognition (CVPR)}, 2009.

\bibitem{sermanet2018time}
P.~Sermanet, C.~Lynch, Y.~Chebotar, J.~Hsu, E.~Jang, S.~Schaal, S.~Levine, and
  G.~Brain.
\newblock Time-contrastive networks: Self-supervised learning from video.
\newblock In {\em 2018 IEEE International Conference on Robotics and Automation
  (ICRA)}, pages 1134--1141. IEEE, 2018.

\bibitem{simoncelli199914}
E.~P. Simoncelli.
\newblock Bayesian multi-scale differential optical flow.
\newblock 1999.

\bibitem{simonyan2014two}
K.~Simonyan and A.~Zisserman.
\newblock Two-stream convolutional networks for action recognition in videos.
\newblock In {\em Advances in neural information processing systems}, pages
  568--576, 2014.

\bibitem{skafte2018deep}
N.~Skafte~Detlefsen, O.~Freifeld, and S.~Hauberg.
\newblock Deep diffeomorphic transformer networks.
\newblock In {\em Proceedings of the IEEE Conference on Computer Vision and
  Pattern Recognition}, pages 4403--4412, 2018.

\bibitem{song2017thin}
J.~Song, L.~Wang, L.~Van~Gool, and O.~Hilliges.
\newblock Thin-slicing network: A deep structured model for pose estimation in
  videos.
\newblock In {\em Proceedings of the IEEE Conference on Computer Vision and
  Pattern Recognition}, pages 4220--4229, 2017.

\bibitem{szeliski2012bayesian}
R.~Szeliski.
\newblock {\em Bayesian modeling of uncertainty in low-level vision},
  volume~79.
\newblock Springer Science \& Business Media, 2012.

\bibitem{trottenberg2000multigrid}
U.~Trottenberg, C.~W. Oosterlee, and A.~Schuller.
\newblock {\em Multigrid}.
\newblock Elsevier, 2000.

\bibitem{tung2017self}
H.-Y. Tung, H.-W. Tung, E.~Yumer, and K.~Fragkiadaki.
\newblock Self-supervised learning of motion capture.
\newblock In {\em Advances in Neural Information Processing Systems}, pages
  5236--5246, 2017.

\bibitem{vijayanarasimhan2017sfm}
S.~Vijayanarasimhan, S.~Ricco, C.~Schmid, R.~Sukthankar, and K.~Fragkiadaki.
\newblock Sfm-net: Learning of structure and motion from video.
\newblock {\em arXiv preprint arXiv:1704.07804}, 2017.

\bibitem{von2007predictive}
C.~Von~Hofsten, O.~Kochukhova, and K.~Rosander.
\newblock Predictive tracking over occlusions by 4-month-old infants.
\newblock {\em Developmental Science}, 10(5):625--640, 2007.

\bibitem{vondrick2016generating}
C.~Vondrick, H.~Pirsiavash, and A.~Torralba.
\newblock Generating videos with scene dynamics.
\newblock In {\em Advances In Neural Information Processing Systems}, pages
  613--621, 2016.

\bibitem{vondrick2018tracking}
C.~Vondrick, A.~Shrivastava, A.~Fathi, S.~Guadarrama, and K.~Murphy.
\newblock Tracking emerges by colorizing videos.
\newblock In {\em Proceedings of the European Conference on Computer Vision
  (ECCV)}, pages 391--408, 2018.

\bibitem{wang2015unsupervised}
X.~Wang and A.~Gupta.
\newblock Unsupervised learning of visual representations using videos.
\newblock In {\em Proceedings of the IEEE International Conference on Computer
  Vision}, pages 2794--2802, 2015.

\bibitem{wang2017transitive}
X.~Wang, K.~He, and A.~Gupta.
\newblock Transitive invariance for self-supervised visual representation
  learning.
\newblock In {\em Proceedings of the IEEE International Conference on Computer
  Vision}, pages 1329--1338, 2017.

\bibitem{xiaolong2019learning}
X.~Wang, A.~Jabri, and A.~A. Efros.
\newblock Learning correspondence from the cycle-consistency of time.
\newblock {\em arXiv preprint arXiv:1903.07593}, 2019.

\bibitem{wang2018occlusion}
Y.~Wang, Y.~Yang, Z.~Yang, L.~Zhao, P.~Wang, and W.~Xu.
\newblock Occlusion aware unsupervised learning of optical flow.
\newblock In {\em Proceedings of the IEEE Conference on Computer Vision and
  Pattern Recognition}, pages 4884--4893, 2018.

\bibitem{wei2018learning}
D.~Wei, J.~J. Lim, A.~Zisserman, and W.~T. Freeman.
\newblock Learning and using the arrow of time.
\newblock In {\em Proceedings of the IEEE Conference on Computer Vision and
  Pattern Recognition}, pages 8052--8060, 2018.

\bibitem{wu2016physics}
J.~Wu, J.~J. Lim, H.~Zhang, J.~B. Tenenbaum, and W.~T. Freeman.
\newblock Physics 101: Learning physical object properties from unlabeled
  videos.
\newblock In {\em BMVC}, volume~2, page~7, 2016.

\bibitem{xie2018object}
C.~Xie, Y.~Xiang, D.~Fox, and Z.~Harchaoui.
\newblock Object discovery in videos as foreground motion clustering.
\newblock {\em arXiv preprint arXiv:1812.02772}, 2018.

\bibitem{yang2018efficient}
L.~Yang, Y.~Wang, X.~Xiong, J.~Yang, and A.~K. Katsaggelos.
\newblock Efficient video object segmentation via network modulation.
\newblock In {\em Proceedings of the IEEE Conference on Computer Vision and
  Pattern Recognition}, pages 6499--6507, 2018.

\bibitem{yang2013articulated}
Y.~Yang and D.~Ramanan.
\newblock Articulated human detection with flexible mixtures of parts.
\newblock {\em IEEE transactions on pattern analysis and machine intelligence},
  35(12):2878--2890, 2013.

\bibitem{yang2018every}
Z.~Yang, P.~Wang, Y.~Wang, W.~Xu, and R.~Nevatia.
\newblock Every pixel counts: Unsupervised geometry learning with holistic 3d
  motion understanding.
\newblock In {\em European Conference on Computer Vision}, pages 691--709.
  Springer, 2018.

\bibitem{yu2015direct}
R.~Yu, C.~Russell, N.~D. Campbell, and L.~Agapito.
\newblock Direct, dense, and deformable: Template-based non-rigid 3d
  reconstruction from rgb video.
\newblock In {\em Proceedings of the IEEE International Conference on Computer
  Vision}, pages 918--926, 2015.

\bibitem{zhan2018unsupervised}
H.~Zhan, R.~Garg, C.~Saroj~Weerasekera, K.~Li, H.~Agarwal, and I.~Reid.
\newblock Unsupervised learning of monocular depth estimation and visual
  odometry with deep feature reconstruction.
\newblock In {\em Proceedings of the IEEE Conference on Computer Vision and
  Pattern Recognition}, pages 340--349, 2018.

\bibitem{zhang2017split}
R.~Zhang, P.~Isola, and A.~A. Efros.
\newblock Split-brain autoencoders: Unsupervised learning by cross-channel
  prediction.
\newblock In {\em Proceedings of the IEEE Conference on Computer Vision and
  Pattern Recognition}, pages 1058--1067, 2017.

\bibitem{zhou2017unsupervised}
T.~Zhou, M.~Brown, N.~Snavely, and D.~G. Lowe.
\newblock Unsupervised learning of depth and ego-motion from video.
\newblock In {\em Proceedings of the IEEE Conference on Computer Vision and
  Pattern Recognition}, pages 1851--1858, 2017.

\bibitem{zhou2016learning}
T.~Zhou, P.~Krahenbuhl, M.~Aubry, Q.~Huang, and A.~A. Efros.
\newblock Learning dense correspondence via 3d-guided cycle consistency.
\newblock In {\em Proceedings of the IEEE Conference on Computer Vision and
  Pattern Recognition}, pages 117--126, 2016.

\bibitem{zhou2016view}
T.~Zhou, S.~Tulsiani, W.~Sun, J.~Malik, and A.~A. Efros.
\newblock View synthesis by appearance flow.
\newblock In {\em European conference on computer vision}, pages 286--301.
  Springer, 2016.

\bibitem{zoph2016neural}
B.~Zoph and Q.~V. Le.
\newblock Neural architecture search with reinforcement learning.
\newblock {\em arXiv preprint arXiv:1611.01578}, 2016.

\end{thebibliography}
}

\newpage\newpage

\setcounter{section}{0}

\clearpage\mbox{}

\begin{center}
 {\large \textbf{Appendix}}
\end{center}

\emph{
In the appendix,
we first show all intermediate results of multigrid Predictive Filter Flow (mgPFF)
from multi-resolution inputs, to have an idea how these outputs look like in terms of
frame reconstruction.
Then, we plot the graph visualization of our model architecture with detailed design of
the two stream architecture.
Furthermore, we visualize pixel embedding generated by our architecture to understand
what the model learns.
Finally, along this document,
we provide some demo videos of the object segmentation/tracking results with different
setup.
}

\section{Intermediate Reconstruction by mgPFF}

\begin{figure*}[t]
    \centering
    \begin{minipage}{0.98\textwidth}
        \centering
        \includegraphics[width=1\linewidth]{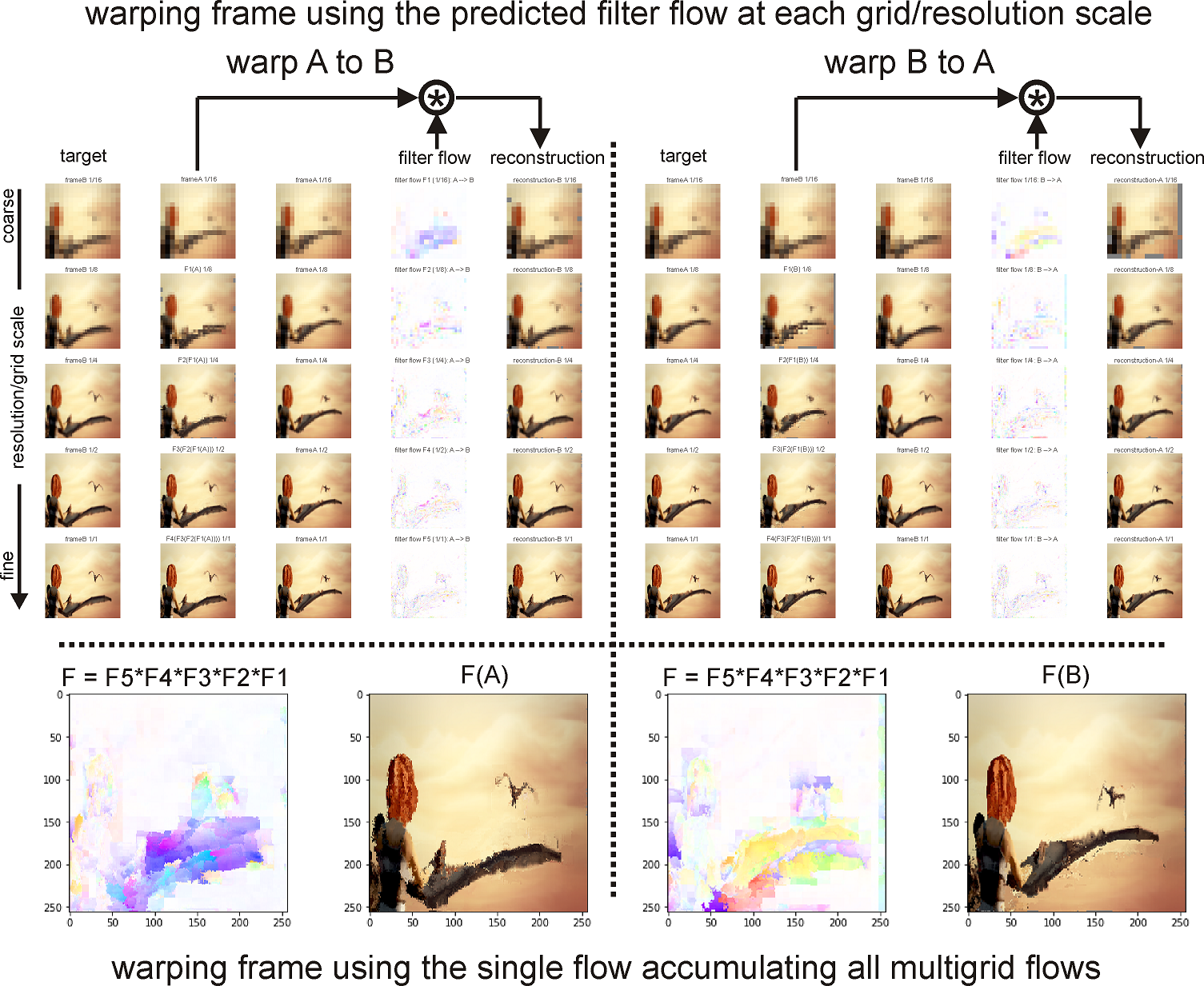}
        \captionsetup{width=0.999\textwidth}
    \end{minipage}
    \caption{\small {\bf Visualization of intermediate results at each resolution scale (grid)}.
     Top: we show the predicted filter flows and the reconstruction results from warping A to B,
      or B to A.
      Note how mgPFF resolves the aliasing effect reflected by the blocks in the
      reconstruction images.
     Bottom: we accumulate all the filter flows (with necessary upsampling using nearest neighbor
     interpolation),
     and transform into a coordinate flow which can be thought as optical flow.
     Then we use the overall flow to warp from one frame to the other.
     This introduces some artifacts due to information loss,
     but the reconstruction appears good generally, e.g., capturing the bird wings' movement.
     In our experiment of tracking,
     we use the coordinate flow in the same way to warp the given masks
     (or the predicted mask at previous frames) to propagate the track results.
    }
    \label{fig:mgPFF_allresults}
\end{figure*}

As our mgPFF performs progressively from coarse to fine,
it produces the predicted filter flows and reconstruction frames at
each resolution scale.
We visualize all the intermediate results in Fig.~\ref{fig:mgPFF_allresults}.
We also accumulate the filter flow maps at all scales and convert it into
the coordinate flow, which can be thought as optical flow.
We use this coordinate flow to warp masks for propagating the track results
in our experiments. Please pay attention to how mgPFF achieves excellent
reconstruction results from coarse to fine,
like resolving the aliasing and block effects, refining reconstruction at finer scales, etc.

\section{Graph Visualization of mgPFF architecture}

\begin{figure*}[t]
    \centering
    \begin{minipage}{0.99\textwidth}
        \centering
        \includegraphics[width=1\linewidth]{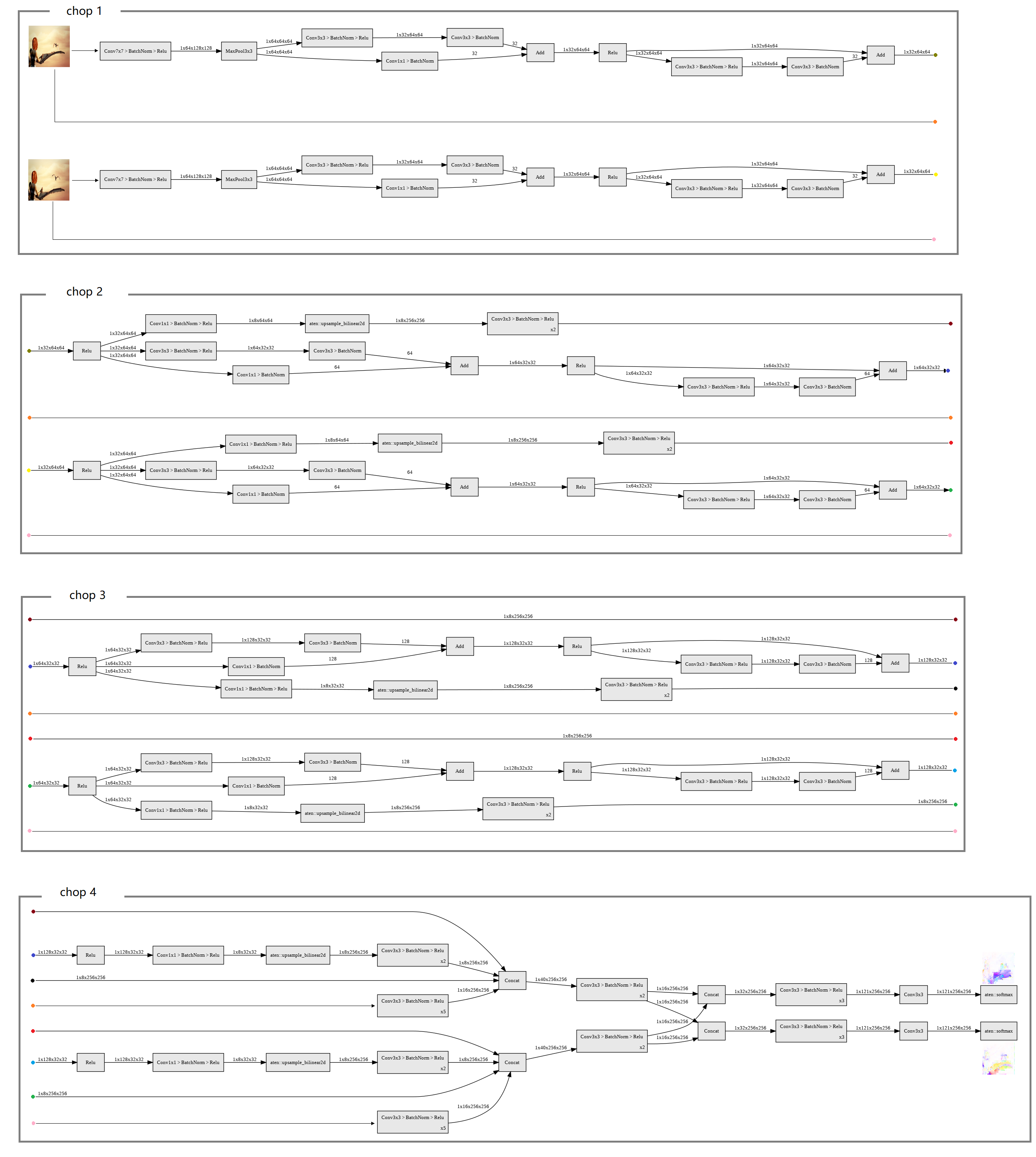}
        \captionsetup{width=0.999\textwidth}
    \end{minipage}
    \caption{\small Graph visualization of mgPFF architecture using
    HiddenLayer toolbox~\cite{Abdulla2018hiddenLayer}.
    Zoom in to see clearly.
    }
    \label{fig:arch}
\end{figure*}

In Fig.~\ref{fig:arch},
we plot the architecture of our model using the HiddenLayer
toolbox~\cite{Abdulla2018hiddenLayer}. As the visualization
is too ``long'' to display, we chop it into four parts.
We modify the ResNet18~\cite{he2016deep}
by removing  res$_4$ and res$_5$ (the top 9 residual blocks,
and reducing the unique channel size
from $[64, 128, 256, 512]$ to $[32, 64, 128, 196]$.
The two macro towers take the two frames,
respectively; in each tower, there are two streams,
one is of U-shape~\cite{ronneberger2015u} with pooling and upsampling
layers to increase the receptive fields, the other is
full-resolution yet shallow in channel depth.
The two-stream architecture is popular in multiple domain learning~\cite{simonyan2014two},
but we note that such design on a single domain was first used in \cite{pohlen2017full}
which is more computationally
expensive that the two streams talk to each other along the whole network flow; whereas ours is cheaper that
they only talk at the top layer.
Our mgPFF is very compact that the overall model size is only \textbf{4.6MB};
it also performs fast that the wall-clock time for processing a pair of 256x256
frames is 0.1 seconds.

As we did not search over architecture design in our work,
tt is worth exploring other sophisticated modules to make it more compact
for deploying in mobile devices, e.g., using meta-learning for architecture search~\cite{zoph2016neural}.

\section{Pixel Embedding in mgPFF }

\begin{figure}[t]
    \centering
    \begin{minipage}{0.44\textwidth}
        \centering
        \includegraphics[width=1\linewidth]{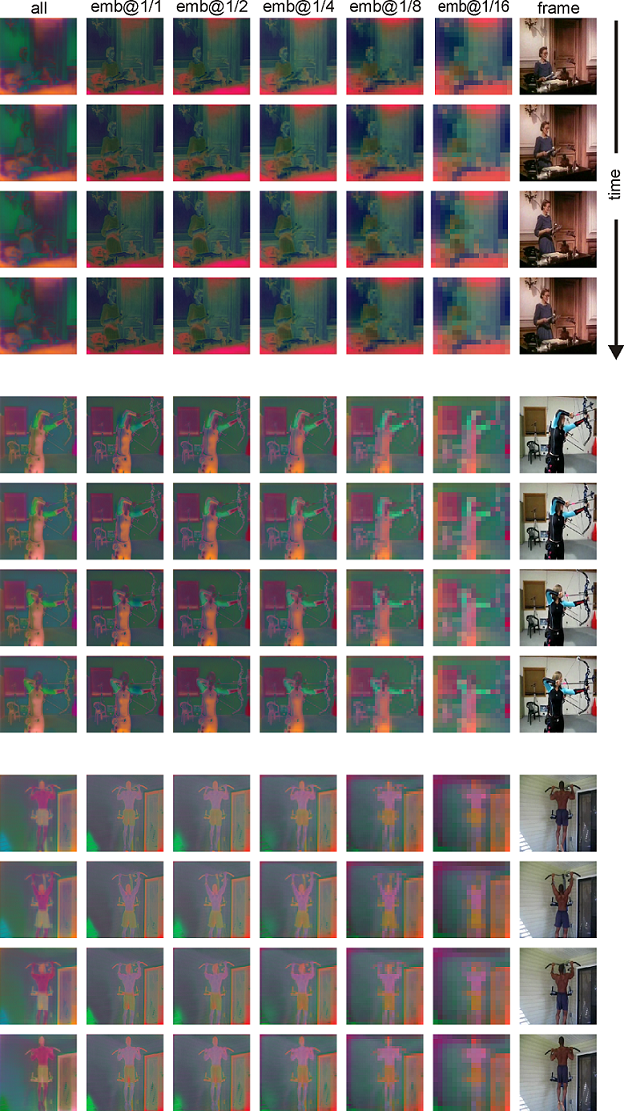}
        \captionsetup{width=0.999\textwidth}
    \end{minipage}
    \caption{\small {\bf Visualization of learned pixel embedding}:
     We use PCA to project the pixel embedding
     (3D array of size H$\times$W$\times$$D$)
     into  H$\times$W$\times$3,
     and visualize it as an RGB image.
     Individual embedding map has $D=16$ in channel dimension.
     We also concatenate the pixel embeddings of all resolutions and
     apply PCA, in which case $D=16*5=80$.
     From the visualization,
     we can see that the visualization colors largely come from the RGB
     intensities. This is largely due to two reasons: 1) the photometric loss we are using
     during training is based on RGB intensities, 2) our mgPFF by nature is based on
     low-level vision that it does not need understanding of mid/high-level perspective of
     the frames.
    }
    \label{fig:embVis}
\end{figure}

As our model produces per-image pixel
embeddings~\cite{newell2017associative, kong2018recurrent}
(the output before ``concatenation layer''
as shown in the architecture Fig.~\ref{fig:arch}),
we are interested in visualizing the pixel embeddings to see what the model learns.
To visualize the pixel embeddings,
we use PCA to project the embedding feature map H$\times$W$\times$D at each resolution/grid
into an H$\times$W$\times$3 array, and visualize the projection as an RGB image.
We also concatenate the embedding maps at all the resolutions/grids
for visualization (with necessary nearest neighbor upsampling).
Fig.~\ref{fig:embVis} lists these visualizations, from which we can see the embedding colors
largely come from the original RGB intensities.
We conjecture this is due to two reasons.
First,
we use a simplistic photometric loss on the RGB values, this explains why the visualization
colors group the pixels which have similar RGB values in local neighborhood.
Second,
our mgPFF by nature is based on low-level vision, i.e., flow field,
and in such a way it does not necessarily depend on mid/high-level understanding of the frames.
Therefore, part/instance grouping does not appear in the embedding visualization,
which is shown in mid-level methods~\cite{vondrick2018tracking, xiaolong2019learning}.
This suggests further exploration of using other losses and combining other mid/high-level
cues to force the model to learn more abstract features.

\section{Video Demos}

\begin{figure*}[t]
    \centering
    \begin{minipage}{0.95\textwidth}
        \centering
        \includegraphics[width=1\linewidth]{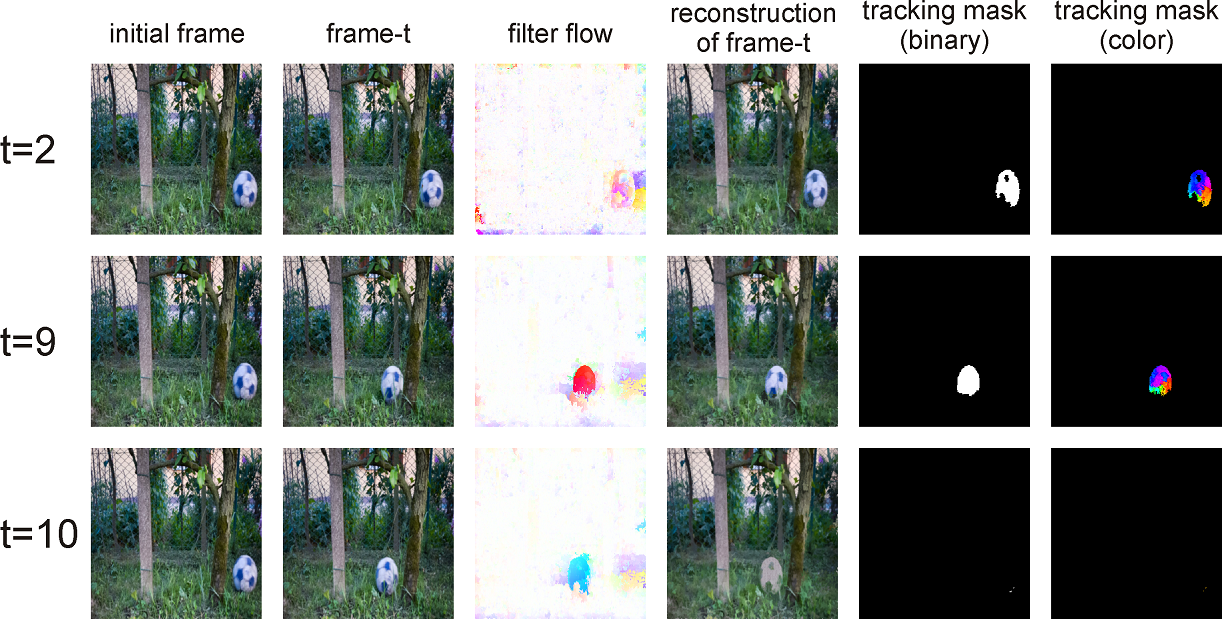}
        \captionsetup{width=0.999\textwidth}
    \end{minipage}
    \caption{\small How far the model can track the object correctly?
    As we adopt the multigrid computing strategy,
    the filter of size 11x11 on the coarsest grid (16x downsample) implies
    the largest displacement we can represent is D=88.
    If the object moves further than D from its last location,
    the model fails in tracking it. This happens at frame-10,
    in which we can see that not only the tracking is missing,
    but also the filter flow changes abruptly and the reconstruction becomes
    very different.
    It turns out that in the reconstruction,
    the soccerball's color is from the grass and tree trunk.
    }
    \label{fig:largestDisplacement}
\end{figure*}

The attached videos demonstrate how mgPFF performs with different setup\footnote{Here is a \href{https://www.youtube.com/playlist?list=PLeUWdu37dSLp68AsgE8RM2x-HJjU_2aEE}{Youtube list}
}.
Note how it improves performance with different setup
in terms of dealing with occlusion and large displacement.

Among the videos, it is worth noting how far the model can go with tracking correctly.
As we adopt the multigrid computing strategy,
the filter of size 11x11 on the coarsest grid (16x downsample) implies
the largest displacement we can represent is D=88.
If we simply warp from the first frame to the $t^{th}$ frame,
it only works well when the total displacement is less that D.
This can be seen from video \underline{\emph{soccerball, $K$=1, frame-$[1]$}} as an example.
When the soccerball moves further than D from its initial location at the first frame,
the model suddenly fails in tracking that the mask is no longer correctly warped.
We show the relevant frames in Fig.~\ref{fig:largestDisplacement}.
It is clear that not only the tracking is missing,
but also the filter flow changes abruptly and the reconstruction becomes
very different.
It turns out that in the reconstruction,
the soccerball's color is from the grass and tree trunk.

Here is the list of videos with brief description:
\begin{enumerate}
  \item \underline{\emph{ soccerball, $K$=3, frame-$[1, t-2, t-1]$}}:
    this video shows the results on soccerball from DAVIS dataset when we feed the \textbf{first frame-1 and two previous frame ($t-2$ and $t-1$)} to predict the filter flow,
    warp frame and track the object at current frame-$t$.
    (video url \url{https://youtu.be/M49nLtT1UmY}).
  \item \underline{\emph{ soccerball, $K$=3, frame-$[t-3, t-2, t-1]$}}:
    this video shows the results on soccerball from DAVIS dataset when we feed the \textbf{previous three frames
     ($t-3$, $t-2$ and $t-1$)} to predict the filter flow, warp frame and track the object
     at current frame-$t$ (video url \url{https://youtu.be/q_FNk-3lh3g}).
  \item \underline{\emph{soccerball, $K$=2, frame-$[1, t-1]$}},
    this video shows the results on soccerball from DAVIS dataset when we feed the \textbf{first frame-1 and one previous frame-($t-1$)} to predict the filter flow, warp frame and track the object.
    (video url \url{https://youtu.be/u6IdVS2L7-M}).
  \item \underline{\emph{soccerball, $K$=1, frame-$[1]$}},
    this video shows the results on soccerball from DAVIS dataset when we feed the \textbf{first frame only }at which the mask is given to predict the filter flow, warp frame and track the object.
    (video url \url{https://youtu.be/vsXZgdR4XEY})
  \item \underline{\emph{soccerball, $K$=1, frame-$[t-1]$}},
    this video shows the results on soccerball from DAVIS dataset when we feed the \textbf{the previous frame-$(t-1)$} to predict the filter flow, warp frame and track the object
    at current frame-$t$.
    (video url \url{https://youtu.be/8AZ9wPF15QE})

  \item \underline{\emph{ dog, $K$=3, frame-$[1, t-2, t-1]$}}:
    this video shows the results on dog from DAVIS dataset when we feed the \textbf{first frame-1 and two previous frame ($t-2$ and $t-1$)} to predict the filter flow,
    warp frame and track the object at current frame-$t$.
    (video url \url{https://youtu.be/seg5tFSMFX8}).
  \item \underline{\emph{ dog, $K$=3, frame-$[t-3, t-2, t-1]$}}:
    this video shows the results on dog from DAVIS dataset when we feed the \textbf{previous three frames
     ($t-3$, $t-2$ and $t-1$)} to predict the filter flow, warp frame and track the object
     at current frame-$t$ (video url \url{https://youtu.be/BqM4-OctYwA}).
  \item \underline{\emph{dog, $K$=2, frame-$[1, t-1]$}},
    this video shows the results on dog from DAVIS dataset when we feed the \textbf{first frame-1 and one previous frame-($t-1$)} to predict the filter flow, warp frame and track the object.
    (video url \url{https://youtu.be/dOao8qQMsv0}).
  \item \underline{\emph{dog, $K$=1, frame-$[1]$}},
    this video shows the results on dog from DAVIS dataset when we feed the \textbf{first frame only }at which the mask is given to predict the filter flow, warp frame and track the object.
    (video url  \url{https://youtu.be/xNMuMlcvfJY})
  \item \underline{\emph{dog, $K$=1, frame-$[t-1]$}},
    this video shows the results on dog from DAVIS dataset when we feed the \textbf{the previous frame-$(t-1)$} to predict the filter flow, warp frame and track the object
    at current frame-$t$.
    (video url \url{https://youtu.be/Yu5amZf1KEc})

\end{enumerate}

\end{document}